%% file: tacl2021v1-template.tex
\documentclass[11pt,a4paper]{article}
\usepackage{times,latexsym}
\usepackage{url}
\usepackage[T1]{fontenc}

\usepackage[acceptedWithA]{tacl2021v1}

\usepackage{times}
\usepackage{latexsym}
\usepackage{float}
\usepackage[compatibility=false,font=small]{caption}
\usepackage[dvipsnames]{xcolor}
\usepackage{adjustbox}

\usepackage[T1]{fontenc}

\usepackage[utf8]{inputenc}
\usepackage{fix-cm}

\usepackage{microtype}
\usepackage{enumitem}

\setlist{topsep=2pt,itemsep=1pt,parsep=0pt,partopsep=0pt}
\usepackage{inconsolata}

\usepackage{graphicx}
\usepackage{amsmath}
\usepackage{amssymb}
\usepackage{booktabs}
\usepackage{multirow}
\usepackage{array} 

\usepackage{tabularx}

\usepackage{flafter}        
\usepackage{placeins}       
\usepackage{dblfloatfix}    

\usepackage{xspace,mfirstuc,tabulary}

\newif\iftaclinstructions
\taclinstructionsfalse 
\iftaclinstructions
\renewcommand{\confidential}{}
\renewcommand{\anonsubtext}{(No author info supplied here, for consistency with
TACL-submission anonymization requirements)}
\newcommand{\instr}
\fi

\iftaclpubformat 

\else

\fi

\title{When Does Label-Free Deconfounding of Document Embeddings Work?\\[6mm]}

\author{
  Anna Kiepura \\
  ETH Zurich \\
  \texttt{akiepura@ethz.ch}
  \And
  Yingqiang Gao \\
  Department of Computational Linguistics \\
  University of Zurich \\
  \texttt{yingqiang.gao@cl.uzh.ch}
  \AND
  Elliott Ash \\
  ETH Zurich \\
  \texttt{ashe@ethz.ch}
  \And
  Yu Fan \\
  The University of Hong Kong \\
  \texttt{yufn@hku.hk}
}

\date{}
\usepackage[compatibility=false]{caption}

\begin{document}
\maketitle

\begin{abstract}

Document embeddings encode spurious attributes that can distort semantic comparisons; yet supervised erasure requires metadata that is often unavailable. We systematically investigate label-free deconfounding across corpora and embedding models. Our results show that label-free methods improve semantic utility over unprocessed embeddings, while simple geometric approaches recover much of the supervised benefit, sometimes matching metadata-supervised erasure in linear confounder recoverability and semantic utility. We identify two distinct causes of ineffective erasure: poor proxy–confounder alignment and insufficient intervention coverage. Crucially, confounder recoverability and semantic utility are distinct dimensions of deconfounding success: utility can improve despite residual confounder information, while stronger suppression does not necessarily improve utility. Moreover, reduced linear recoverability can coexist with substantial nonlinear recoverability. Together, these findings distinguish confounder discovery from intervention effectiveness and highlight the importance of evaluating nuisance suppression alongside semantic preservation.\footnote{Both code and data are available in our \href{https://github.com/annamkiepura/label-free-deconfounding-embeddings}{GitHub repository}.}
\end{abstract}

\section{Introduction}

Text embeddings underpin semantic search, document retrieval, classification, and clustering \citep{reimers-gurevych-2019-sentence}. These applications rely on a simple assumption: \emph{documents with similar meanings should be close in the embedding space}. Yet embeddings also encode attributes such as language, source, genre, publication venue, writing style, and author identity. In heterogeneous corpora, such attributes can distort embedding comparisons. They may bring unrelated documents closer together or separate documents that describe the same events, arguments, or topics \citep{yang-etal-2021-simple, fan-etal-2025-medium}. Consequently, retrieval and clustering may reflect how a corpus was assembled rather than the semantic relationships it aims to recover.



Concept erasure offers a way to reduce these distortions. When confounder labels are available, supervised erasure can improve semantic comparisons \citep{fan-etal-2025-medium}. In practice, however, the relevant metadata may be unavailable, incomplete, or unreliable. For example, source provenance can be lost as text datasets are compiled and repackaged \citep{longpre2024audit}. This motivates label-free erasure: identifying and suppressing nuisance structure using only pooled embeddings. We study this setting without confounder annotations, textual descriptions, population priors, or a predefined number of confounder classes.

Label-free erasure poses two distinct challenges: \emph{discovery}, which involves identifying confounder-related structures from embedding geometry, and \emph{intervention}, which involves suppressing them without sacrificing semantic utility. Distinguishing these challenges is essential for understanding why erasure succeeds or fails.

We address three main questions: (i) how much of the benefit of metadata-supervised erasure is achievable without labels; (ii) which geometric and intervention factors explain success or failure; and (iii) how confounder suppression relates to semantic utility. We investigate these questions across six corpora and ten embedding models, comparing PCA-, clustering-, and sparse-autoencoder (SAE)-based erasure. Confounder labels are used for dataset construction, supervised references, and evaluation, but never for label-free fitting or selection.

Our results show that label-free deconfounding can recover substantial semantic utility benefits without confounder labels. In our primary evaluation, all three methods improve the mean semantic utility over unprocessed embeddings, while simple geometric approaches sometimes match metadata-supervised erasure despite incomplete confounder removal. However, both the effectiveness of confounder suppression and the relative performance of different methods depend on the geometry of the corpus and the embedding model. Further, we identify poor proxy–confounder alignment and insufficient intervention extent as distinct causes of weak erasure, highlighting that confounder discovery alone does not guaranty effective intervention. Moreover, we show that confounder recoverability and downstream semantic utility are distinct dimensions of deconfounding success: substantial improvements in semantic utility can occur even when confounder information remains recoverable from the embeddings. Additionally, substantial nonlinear confounder recoverability can persist despite large semantic utility gains and near-chance linear recoverability. Together, these findings highlight the importance of distinguishing confounder discovery from intervention effectiveness and evaluating semantic utility alongside both linear and nonlinear confounder recoverability.

Our contributions are:
\begin{itemize}
    \item We demonstrate that label-free deconfounding can recover substantial semantic benefits without nuisance annotations, sometimes approaching metadata-supervised performance.

    \item We identify the geometric conditions and distinct failure modes underlying label-free erasure, providing an explanatory basis for diagnosing success and failure across corpora and embedding models.

    \item We disentangle feature discovery, intervention construction, and erasure extent, showing why more complex methods need not outperform appropriately matched geometric baselines.

    \item We provide a systematic evaluation framework spanning linear and nonlinear confounder leakage, in-distribution semantic utility, and unseen-task transfer, revealing why no single metric establishes deconfounding success.
\end{itemize}

\section{Related Work}

Prior work shows that dense text embeddings encode nuisance attributes alongside semantic information, potentially distorting downstream comparisons. Language-specific structure in multilingual embeddings \citep{libovicky-etal-2020-language,tiyajamorn-etal-2021-language} contributes to same-language retrieval biases \citep{roy-etal-2020-lareqa,yang-etal-2021-simple} and inconsistent rankings for semantically equivalent queries \citep{yang-etal-2024-language-bias}. Document-source information likewise affects similarity and clustering \citep{fan-etal-2025-medium}, while neural retrievers favor LLM-generated over comparable human-written text \citep{dai-etal-2024-neural-retrievers}, with evidence of corresponding source-associated directions in embedding space \citep{huang-etal-2026-data-not-model}.

Unwanted attributes can be suppressed during training through adversarial objectives \citep{elazar-goldberg-2018-adversarial} or erasure-based regularization \citep{huang-etal-2024-language}. Post-hoc concept erasure removes information associated with specified attributes through methods including concept-subspace projection \citep{bolukbasi2016man}, iterative nullspace projection (INLP) \citep{ravfogel-etal-2020-null}, and adversarial linear erasure (R-LACE) \citep{pmlr-v162-ravfogel22a}. LEACE provides a closed-form affine transformation that removes linearly accessible concept information while minimizing representation distortion \citep{belrose2025leaceperfectlinearconcept}. Applied to frozen document embeddings, metadata-supervised LEACE improves retrieval and clustering \citep{fan-etal-2025-medium}. Other approaches relax instance-level confounder labels but require population-level attribute information \citep{shao-etal-2023-erasure}, downstream task labels \citep{masoudian-etal-2024-unlabeled}, or natural-language confounder descriptions \citep{gao-etal-2026-mitigating}. These studies establish the benefits of targeted erasure under varying supervision requirements but leave open whether nuisance structure can be discovered and suppressed from pooled embeddings alone.

Unsupervised geometric post-processing has improved sentence embeddings through the removal of leading principal components \citep{arora2017simple}, and word embeddings through the removal or attenuation of high-variance directions \citep{mu2018all,liu2019unsupervised}. However, these methods target general representation quality rather than identifying and suppressing specific nuisance attributes. We instead investigate whether embedding geometry provides label-free proxies for language and source confounders, distinguishing proxy--confounder alignment from intervention effectiveness and semantic costs.

Sparse autoencoders (SAEs) learn dictionaries of sparsely activated features, with TopK variants explicitly controlling activation sparsity \citep{gao2025scaling}. Applied to internal language-model activations, they reveal interpretable features and support targeted interventions \citep{huben2024sparse,marks2025sparse}. Applied directly to dense text embeddings, SAEs enable feature interpretation, retrieval steering, and sparse retrieval representations \citep{oneill2024disentangling,kang-etal-2025-interpret,park-etal-2025-decoding}. Related work uses SAE features derived from language-model activations to analyze document properties and dataset correlations \citep{pmlr-v306-jiang26ar}. More directly relevant, SHIFT removes spurious classifier signals through human-guided feature ablation \citep{marks2025sparse}, while SAEBench evaluates spurious-correlation removal using labeled probes \citep{pmlr-v267-karvonen25a}. LangSAE suppresses language-associated features in pooled embeddings but requires language labels for selection and editing \citep{kim2026langsae}. While these studies explore particular feature-selection or editing strategies, they do not jointly disentangle latent selection, erasure-map construction, and intervention extent in the strictly label-free document-embedding setting. We address this gap by systematically comparing alternative latent-selection criteria and erasure operators under matched controls, including displacement-matched geometric baselines, while evaluating both confounder recoverability and semantic utility.

\section{Methodology}
\label{sec:methodology}

\input{tables_4/table_1_datasets}

\subsection{Setting, Corpora, and Embedding Models}



We formalize the document-embedding deconfounding problem studied in this paper as follows. Given a corpus $\mathcal{D}=\{d_i\}_{i=1}^n$, a frozen embedding model $f:\mathcal{D}\to\mathbb{R}^d$ produces row-vector embeddings $x_i=f(d_i)$, stacked in $X\in\mathbb{R}^{n\times d}$. Each document has a confounder attribute $c_i\in\{1,\ldots,C\}$, corresponding to language or document source. We compare affine erasers $T(x)=xA+b$ with $A\in\mathbb{R}^{d\times d}$ and $b\in\mathbb{R}^d$, which aim to reduce the recoverability of $c_i$ from $T(x_i)$ while preserving semantic utility. The supervised reference uses metadata-derived labels $c_i$ to fit the eraser. In the label-free setting, we instead seek to infer proxies for $c_i$ from the geometry of pooled training embeddings alone, without using metadata-derived confounder labels at any stage of fitting or selection.

\paragraph{Corpora and composition.} Table~\ref{tab:datasets} summarizes the six corpora, their confounders, and associated semantic-utility evaluations. For SwiLTra \citep{niklaus2025swiltra} and SCOTUS \citep{fang2023superscotus}, we consider the three-class setting and all three pairwise binary subsets; Fedlex is evaluated in the trilingual DE--FR--IT setting. Because the binary and multiclass settings differ in both corpus composition and retrieval index sets, comparisons across them do not isolate the effect of class count. We therefore analyze rank effects within each setting. Dataset construction details are provided in Appendix~\ref{sec:appendix_A1}.

\paragraph{Embedding models.} Multilingual E5-small \citep{wang2024multilingual} is the primary model; MiniLM \citep{reimers-gurevych-2019-sentence} provides targeted SCOTUS and frozen-transfer comparisons. A separate ten-model study refits every eraser in each representation space. Configurations and coverage appear in Appendix~\ref{sec:appendix_A2}.

\subsection{Geometric Erasure and Supervised Reference}
\label{sec:baselines}

Raw denotes the original, untransformed embeddings and serves as the no-erasure reference. LEACE fits an affine transformation that removes linear information about a supplied concept matrix \citep{belrose2025leaceperfectlinearconcept}. In our label-free variants, this concept is derived solely from the geometry of the training embeddings without using confounder labels for fitting or selection.

\paragraph{Metadata-LEACE.} Following \citet{fan-etal-2025-medium}, we construct the concept matrix from one-hot confounder labels obtained from metadata, such as language or document source. This provides a supervised reference with an erasure rank of at most $C-1$.

\paragraph{$k$-means LEACE.} We cluster training embeddings with $k$-means and use one-hot assignments as pseudo-labels for LEACE. We select $K$ by maximizing the training silhouette score \citep{rousseeuw1987silhouettes}, without knowing the true confounder class count $C$; the erasure rank is at most $K-1$. The method differs from Metadata-LEACE only in its concept labels. If clustering recovers the true confounder partition up to permutation, both yield the same transformation (up to numerical precision). In this case, matching supervised performance reflects successful unsupervised confounder discovery rather than a different erasure operator.

\paragraph{PCA-$r$ removal.} As a simpler label-free geometric baseline, we remove the leading $r$ principal directions from the training embeddings. Let $\mu$ denote the training mean, and let $B_r$ contain the leading $r$ orthonormal principal directions as rows. We apply
\begin{equation}
T_{\mathrm{PCA},r}(x)=\mu+(x-\mu)(I-B_r^\top B_r).
\end{equation}
PCA-1 is the primary configuration. 

Numerical settings and implementation details are provided in Appendix~\ref{sec:appendix_geometric}.

\subsection{SAE-Based Erasure}
\label{sec:sae-leace}

PCA and $k$-means provide relatively coarse geometric proxies based on leading variance directions or a single partition of the embedding space. We therefore additionally consider sparse autoencoders (SAEs), which learn an overcomplete dictionary of sparsely activated directions and may expose confounder-related structures not captured by these simpler proxies. We fit a TopK SAE \citep{gao2025scaling} on $X_{\mathrm{tr}}$, yielding decoder directions $W_{\mathrm{dec}}\in\mathbb{R}^{m\times d}$ and sparse activations $F_{\mathrm{tr}}$. Latents are ranked by the empirical mutual information (MI) between their binary firing indicators and the $k$-means assignments defined above. We select the top $N$ latents, with index set $S$ and corresponding decoder rows $D_S\in\mathbb{R}^{N\times d}$. The primary configuration uses $m=8d$, activation sparsity $k=32$, and selection budget $N=32$; $k$ determines how many latents are active per document, whereas $N$ determines how many directions are selected for erasure. We additionally vary $N$ in the erasure-rank sweep (Figure~\ref{fig:fig_1}) and vary both $m$ and $k$ in the SAE-construction control (Appendix Table~\ref{tab:sae-construction}, Appendix~\ref{sec:appendix_construction}). 

Our primary SAE-based intervention, \emph{SAE-LEACE} (\emph{Directions-LEACE}), supplies the coordinates of the training embeddings along the selected decoder directions to LEACE:
\begin{equation}
Z_{\mathrm{tr}}=X_{\mathrm{tr}}D_S^\top.
\end{equation}
The selected directions are treated as candidate axes of confounder-related variation; neither they nor the $k$-means assignments are assumed to recover the true confounder. Residual confounder information is instead evaluated on held-out documents (Section~\ref{sec:evaluation_protocol}). 

SAE training, firing-frequency handling, MI estimation, and latent scoring are detailed in Appendix~\ref{sec:appendix_A3}; numerical settings are given in Table~\ref{tab:hyperparameters}.

\section{Evaluation}
\label{sec:evaluation}

\subsection{Controlled Comparisons}
\label{sec:controls}

For the SAE-based method, we isolate three factors: candidate-direction selection, intervention construction, and erasure extent. Matched conditions share data splits, evaluation folds, and, where applicable, the fitted SAE.

\paragraph{Candidate-direction selection.} Our primary method ranks SAE latents by MI with the $k$-means pseudo-groups (Section~\ref{sec:sae-leace}). We compare this with $k$-means-independent selection rules—random selection, firing entropy, Hartigan's dip statistic, and co-firing scores—as well as a post-hoc true-label oracle and random non-SAE directions. Definitions appear in Appendix~\ref{sec:appendix_selection}; matched results are reported in Table~\ref{tab:mechanism-controls}.

\paragraph{Intervention construction.} Holding the selected latent set $S$ fixed, we compare alternative ways to convert it into an erasure map: Directions-LEACE, Activation-LEACE, and orthogonal projection controls. This isolates the effect of the intervention operator from latent selection. Formal definitions are given in Appendix~\ref{sec:appendix_operators}.

\paragraph{Erasure extent.} To better distinguish direction selection from simply broader intervention, we vary PCA rank $r$, $k$-means cluster count $K$, and SAE budget $N$, and relate them to the effective rank of the fitted map. We also compare SAE-LEACE with PCA matched to its normalized squared training displacement. Definitions and matching details are provided in Appendix~\ref{sec:appendix_extent}; results appear in Figure~\ref{fig:fig_1} and Table~\ref{tab:erasure-extent}.

\subsection{In-Distribution Evaluation}
\label{sec:evaluation_protocol}

\paragraph{Fitting and evaluation.}
For each outer split, all components that determine an eraser are learned exclusively from the training portion: PCA directions; $k$-means assignments and silhouette-based $K$ selection; SAE parameters; latent ranking and selection; and the fitted LEACE maps. The resulting transformations are then frozen. Evaluation uses held-out documents to measure residual confounder leakage and semantic utility; leakage probes are trained and tested by grouped cross-validation entirely within this held-out portion. Corpus-specific outer splits and grouping constraints are defined in Appendix~\ref{sec:appendix_id}.

\paragraph{Confounder leakage.}
To assess whether erasure actually suppresses confounder information, we test how well language or source can still be recovered from transformed held-out embeddings. We fit fresh linear logistic-regression and nonlinear MLP probes and report ROC-AUC, with lower values indicating less recoverable confounder information. Metadata-LEACE serves as the supervised reference; because LEACE guaranties erasure only for the moments on which it is fitted, its held-out AUC need not reach chance. Probe fitting, cross-validation, and the corresponding training-moment sanity check are detailed in Appendix~\ref{sec:appendix_id}.

\paragraph{Semantic utility.}
Because suppressing confounder information may also remove task-relevant semantic structure, we evaluate whether transformed embeddings preserve or improve meaningful document relationships. SemEval, SwiLTra, Fedlex, and SCOTUS use cosine R@1/10 for held-out linked documents against the full transformed corpus, including training documents and unpaired distractors. SemEval additionally uses Spearman correlation with human similarity ratings, while CAP uses cluster-size-weighted topic purity \citep{manning2008introduction} and adjusted Rand index (ARI; \citealp{hubert1985comparing}). Bills–Tweets has no semantic-utility annotations and is evaluated only for leakage. Ranking, clustering, and aggregation details are provided in Appendix~\ref{sec:appendix_id}.

\subsection{Post-Hoc Geometry Diagnostics}
\label{sec:geometry-diagnostics}

To understand \emph{why} a label-free eraser succeeds or fails, we use the true confounder labels only after fitting to diagnose how the learned geometric proxy relates to the confounder-bearing structure. These diagnostics are explanatory only and never enter fitting or model selection. In particular, they help distinguish two failure modes: insufficient dimensional coverage by the intervention and poor alignment between the unsupervised proxy and the true confounder. For PCA, we measure the fraction of the linear confounder cross-covariance subspace captured by the leading principal components; for $k$-means, we measure the ARI between silhouette-selected clusters and the true confounder classes. We relate both diagnostics to each method's held-out linear-AUC gap from Metadata-LEACE (Figure~\ref{fig:fig_2}). Formal definitions and evaluated cohorts are provided in Appendix~\ref{sec:appendix_geometry}.



\subsection{Frozen Out-of-Distribution Transfer}
\label{sec:ood-transfer}

In-distribution improvements do not establish whether corpus-specific erasers preserve general-purpose semantic utility. Since embedding models are often deployed across domains, we evaluate frozen out-of-distribution (OOD) transfer to test whether OOD performance is preserved without significant degradation. Maps fitted on CAP, SCOTUS, or SwiLTra are applied unchanged within the same embedding-model space to five MTEB legal retrieval tasks \citep{muennighoff-etal-2023-mteb} and STSBenchmark \citep{cer-etal-2017-semeval}, without target-specific refitting. We report seed-averaged changes relative to Raw in equal-weight legal macro nDCG@10 and STS Spearman correlation, with paired 95\% bootstrap intervals. Target datasets, donor settings, transferred maps, resampling, and prompt-robustness analyzes are detailed in Appendix~\ref{sec:appendix_A5}.

\section{Results and Discussion}
\label{sec:results}

\subsection{When does label-free erasure approach supervised erasure?}
\label{sec:results-primary}

\input{tables_4/table_2_primary_id}

\paragraph{Label-free erasure can improve semantic utility despite residual leakage.} In the primary E5-small evaluation, PCA-1, \(k\)-means LEACE, and SAE-LEACE all achieve higher mean semantic utility than Raw across the five corpora with utility annotations (Table~\ref{tab:primary-id}). The additional E5-small retrieval, graded similarity, and topic clustering evaluations show the same pattern (Appendix Table~\ref{tab:extended-id}). These gains coexist with residual confounder recoverability, demonstrating that improvements in the evaluated semantic tasks do not require complete nuisance removal.

\paragraph{Geometric proxies can match supervised erasure.} In some settings, label-free erasure reproduces the supervised reference, most clearly on the two trilingual Swiss corpora. On SwiLTra, $k$-means LEACE matches Metadata-LEACE at reported precision in both linear leakage (AUC 0.577) and retrieval (R@10 0.969); Fedlex shows the same pattern (AUC 0.644; R@10 0.981) (Table~\ref{tab:primary-id}). Because the methods use the same LEACE estimator and differ only in their concept labels, this agreement indicates that silhouette-selected $k$-means can recover confounder-aligned partitions without metadata or knowledge of the true class count. The result therefore reflects successful unsupervised nuisance discovery, not a different erasure operator.

\paragraph{Success depends on more than confounder cardinality.} This success does not extend uniformly to other corpora with similar numbers of classes. On the three-source SCOTUS, $k$-means LEACE leaves substantially more residual leakage than Metadata-LEACE (0.699 vs. 0.615), despite similar retrieval (0.787 vs. 0.800) (Table~\ref{tab:primary-id}). On SemEval and CAP, PCA-1 and $k$-means LEACE also remain well above the supervised reference. These contrasts suggest that success depends not merely on the confounder class count but on how well geometric proxies recover nuisance-related structure (Appendix Table~\ref{tab:extended-id}).



\paragraph{Leakage and semantic utility are distinct dimensions of deconfounding success.} On CAP, SAE-LEACE achieves a lower linear AUC but also lower topic purity than PCA-1 and \(k\)-means LEACE (Table~\ref{tab:primary-id}). Stronger suppression, therefore, need not imply greater semantic utility; both outcomes must be measured separately and interpreted jointly. The following analyzes examine intervention coverage and proxy–confounder alignment to explain variation in erasure outcomes.

\subsection{How does erasure rank affect leakage and utility?}
\label{sec:results-rank}

\begin{figure*}[t]
\centering
\includegraphics[width=\textwidth]{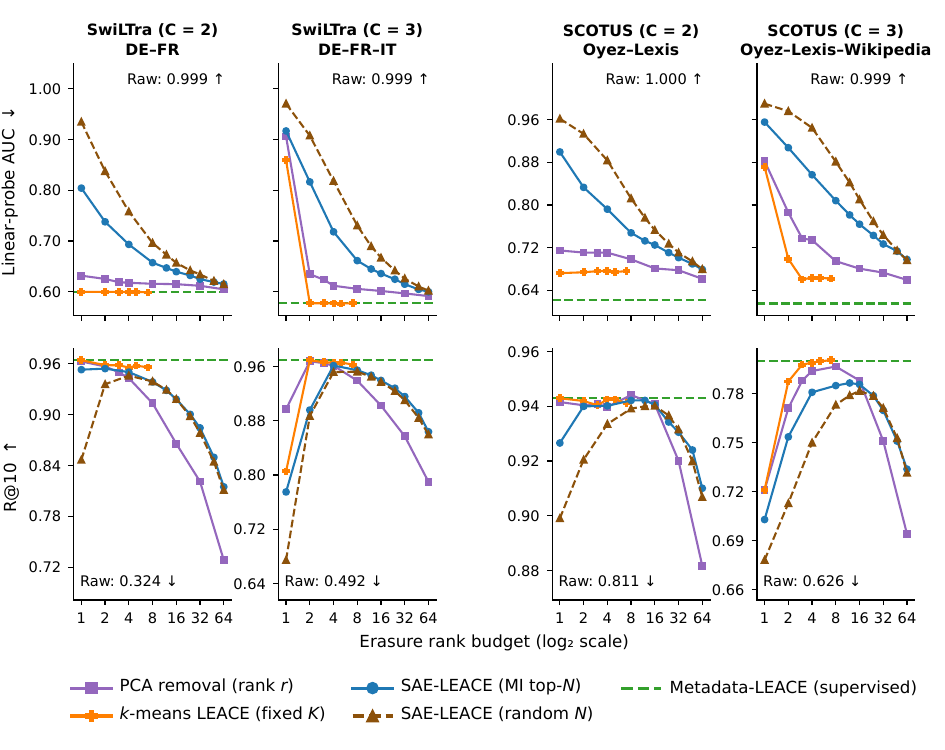}
\vspace{-3pt}
\captionsetup{skip=0pt}
\caption{\textbf{Leakage--utility trade-offs across erasure ranks.}
Held-out linear AUC (top) and retrieval R@10 (bottom) for binary and three-class SwiLTra and SCOTUS with E5-small (means over seeds 0--4). Erasure rank is $r$ for PCA, $K-1$ for $k$-means LEACE, and $N$ for SAE-LEACE; these equal the numerical rank of $I-A$ in all checked fits (Appendix~\ref{sec:appendix_A4}). Random SAE curves average five selections from the same eligible latent pool as MI selection (Appendix~\ref{sec:appendix_A3}).}
\vspace{-10pt}
\label{fig:fig_1}
\end{figure*}

\paragraph{Insufficient rank can limit erasure.} Rank sweeps (Figure~\ref{fig:fig_1}) test whether weak low-rank erasure reflects insufficient intervention coverage. On trilingual SwiLTra, increasing PCA rank from one to two reduces linear AUC from 0.906 to 0.634 and improves R@10 from 0.897 to 0.968. Thus, poor PCA-1 performance need not imply that nuisance structure is inaccessible to geometric removal. Increasing $k$-means from $K=2$ to $K=3$ likewise matches Metadata-LEACE (AUC 0.577; R@10 0.969), although this changes both the rank and the partition.

\paragraph{Higher rank does not guarantee supervised-level erasure.} On SCOTUS, increasing PCA or SAE rank reduces leakage but leaves a substantial gap to Metadata-LEACE (Figure~\ref{fig:fig_1}). Thus, increasing the extent of intervention is not always sufficient, and rank sweeps alone cannot distinguish limited coverage from poor proxy–confounder alignment (Section~\ref{sec:geometry-diagnostics}).

\paragraph{Broader erasure can compromise semantic utility.} On SwiLTra, modest rank increases initially improve both leakage and retrieval, whereas broader PCA and SAE interventions yield diminishing leakage reductions at the cost of retrieval performance (Figure~\ref{fig:fig_1}). MI-selected SAE directions outperform random selections most clearly at smaller budgets, with the advantage narrowing as $N$ increases. Thus, direction selection matters most under constrained budgets, while increasing intervention extent can trade semantic utility for further confounder suppression.

\subsection{When does embedding geometry support label-free erasure?}
\label{sec:results-geometry}

\begin{figure*}[t]
\centering
\includegraphics[width=\textwidth]{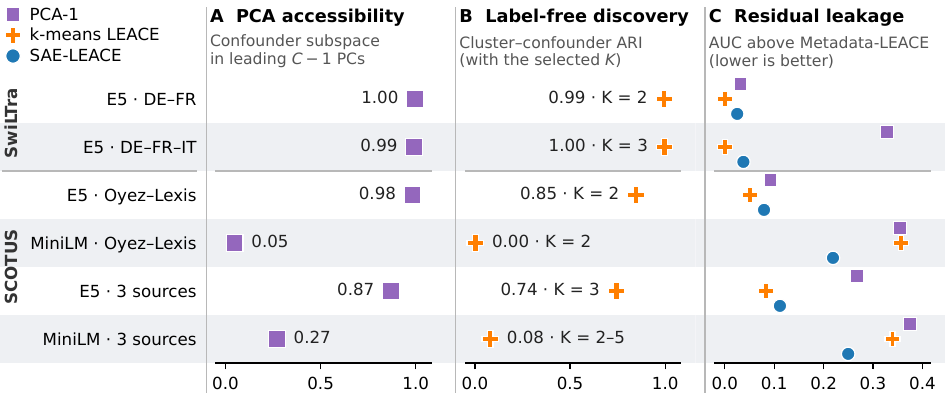}
\vspace{-3pt}
\captionsetup{skip=0pt}
\caption{\textbf{Geometry and nuisance discovery in label-free erasure.} Across six SwiLTra and SCOTUS settings, panels show (A) the fraction of the linear confounder covariance subspace captured by the leading $C-1$ PCs; (B) ARI between silhouette-selected $k$-means groups and confounder classes; and (C) each label-free method's held-out linear-AUC gap to Metadata-LEACE (lower is better; $0$ matches the supervised reference).}
\vspace{-8pt}
\label{fig:fig_2}
\end{figure*}


\paragraph{Weak erasure can reflect insufficient coverage or poor proxy–confounder alignment.} Broader intervention may help when coverage is insufficient, whereas poor alignment motivates revisiting the nuisance proxy. To distinguish these explanations, we relate the rank sweeps (Figure~\ref{fig:fig_1}) and held-out linear-AUC gaps from Metadata-LEACE to two post-hoc geometry diagnostics (Figure~\ref{fig:fig_2}): the fraction of the linear confounder subspace captured by the leading \(C-1\) PCs and the ARI between \(k\)-means clusters and confounder classes.

\paragraph{PCA alignment is model-dependent.} On trilingual SwiLTra, the leading two PCs capture 99\% of the linear confounder subspace; yet PCA-1 leaves substantial leakage (Figure~\ref{fig:fig_2}). Adding the second PC reduces AUC and improves retrieval (Figure~\ref{fig:fig_1}), indicating insufficient coverage rather than poor alignment. On SCOTUS, capture is high with E5-small (0.98 binary; 0.87 three-source) but much lower with MiniLM (0.05 and 0.27). Thus, the accessibility of nuisance structure to low-rank PCA is embedding-model dependent.

\paragraph{Correct cluster count does not guarantee confounder alignment.} Recovering the appropriate number of clusters is insufficient if their assignments do not track the confounder. On binary SCOTUS, both models select $K=2$, but E5-small achieves an ARI of 0.85 and a linear-AUC gap of about 0.05 to Metadata-LEACE, whereas MiniLM has near-zero ARI and a gap of 0.34 (Figure~\ref{fig:fig_2}). On three-source SCOTUS, ARI similarly falls from 0.74 to 0.08, alongside greater residual leakage. Hence, the quality of the recovered partition, not simply its cardinality, determines whether clustering can approximate supervised erasure.

\paragraph{Stronger SAE erasure does not establish better confounder discovery.} SAE-LEACE can reduce leakage where lower-rank proxies fall short, but this does not establish whether it discovers more relevant nuisance structures or simply intervenes more broadly. Figure~\ref{fig:fig_2} diagnoses PCA and $k$-means alignment, whereas the rank sweeps (Figure~\ref{fig:fig_1}) characterize the extent of intervention without directly measuring SAE--confounder alignment. Together, these analyzes show why leakage alone cannot establish proxy quality; the matched controls in Section~\ref{sec:results-mechanism} further separate direction selection, operator choice, and extent of intervention.

\subsection{What makes the SAE-based intervention work?}
\label{sec:results-mechanism}

\input{tables_4/table3_mechanism_controls}

\input{tables_4/table_erasure_extent}

\paragraph{Learned SAE directions are useful, but ranking adds modest gains.} At $N=32$, MI selection modestly reduces linear AUC relative to random SAE directions (0.616 vs. 0.630 on SwiLTra; 0.727 vs. 0.742 on SCOTUS), with little retrieval difference (Table~\ref{tab:mechanism-controls}). Alternative $k$-means-independent rankings do not improve on the matched MI baseline at reported precision, and the true-label MI oracle improves AUC by at most 0.005. Random Gaussian directions perform substantially worse, even after displacement matching. Thus, the learned SAE dictionary supplies useful candidate directions, but their MI-based ranking contributes relatively little at the primary budget. Larger eligible pools yield clearer MI--random gaps, although dictionary size or sparsity also change (Appendix Table~\ref{tab:sae-construction}).


\paragraph{Useful features do not guarantee effective erasure.} We hold the selected latent set (S) fixed to isolate the effect of intervention construction. Activation-LEACE leaves substantially more leakage than Directions-LEACE, while the orthogonal removal of the selected decoder span leaves linear leakage close to Raw (Table~\ref{tab:mechanism-controls}). Orthogonal removal of $\operatorname{span}(\Sigma D_S^\top)$, the Directions-LEACE displacement subspace, matches Directions-LEACE in linear AUC and R@10, with nearly identical MLP AUC. Mean restoration has a negligible effect. Crucially, this shows that the subspace targeted by the intervention, rather than the LEACE operator itself, is central to effective suppression in these settings. These controls identify the induced displacement subspace as central to suppression and show why feature discovery and intervention construction require separate evaluation.


\paragraph{Apparent method advantages can reflect intervention extent. } Comparing SAE-LEACE with PCA-1 alone conflates intervention type with extent. Across all settings in Table~\ref{tab:erasure-extent}, displacement-matched PCA uses ranks 18–23 and achieves a lower mean linear AUC than SAE-LEACE; the paired interval reaches zero only on SemEval. Relative to SAE-LEACE, matched PCA yields lower utility on SwiLTra, CAP, and Fedlex, similar utility on E5-small SCOTUS and SemEval, and higher utility on MiniLM SCOTUS. The leakage pattern persists across the SAE-construction controls, although some paired intervals include zero (Appendix Table~\ref{tab:sae-construction}). Thus, SAE-LEACE does not consistently outperform a comparably broad PCA intervention. Establishing more selective erasure requires assessing leakage and semantic utility jointly while accounting for the extent of the intervention.

\paragraph{Linear erasure does not guarantee complete confounder removal.} Nonlinear leakage remains substantial after both supervised and label-free erasure. Metadata-LEACE is therefore a supervised reference for \emph{linear} erasure, not complete confounder removal. Taken together, these controls limit what can be attributed to SAE-based confounder discovery: while learned directions are useful, their effectiveness depends more strongly on how they are converted into an intervention and how broadly the embedding space is modified.

\subsection{How robust are the findings across embedding models and unseen tasks?}
\label{sec:results-robustness}


\paragraph{Label-free success depends on embedding-model geometry.} The model-refitting study (Appendix Table~\ref{tab:ten-encoder-robustness}) shows that all three label-free methods reduce mean linear confounder recoverability and improve mean semantic utility relative to Raw across all five corpora, although the magnitude varies across encoders. On binary SwiLTra, $k$-means LEACE achieves large, consistent improvements (mean $\Delta$AUC $=-0.372$; $\Delta$R@10 $=+0.235$), alongside near-perfect cluster–language alignment. In contrast, leakage reductions are negligible on SCOTUS for encoders with near-zero cluster–source ARI, while effectiveness varies substantially on SemEval and Fedlex. SAE-LEACE often achieves stronger leakage reductions in challenging settings, but not necessarily greater utility: on CAP, it yields the largest mean AUC reduction ($-0.257$) but almost no utility improvement ($+0.003$). Together with Section~\ref{sec:results-geometry}, these findings show that label-free effectiveness depends on embedding-space geometry, with no method consistently dominating across encoders and metrics.


\paragraph{Semantic preservation requires evaluation beyond the fitting corpus.} We apply maps fitted on CAP, SCOTUS, and SwiLTra unchanged within their respective embedding spaces to five unseen legal retrieval tasks and STSBenchmark (Appendix Figure~\ref{fig:fig_B3}). This measures collateral semantic utility, not target-confounder removal. Low-rank transformations generally have small transfer effects, whereas broader interventions incur larger legal-retrieval losses. Across five donor–model settings, SAE-LEACE reduces legal macro nDCG@10 relative to Raw by 0.007–0.020, comparable to displacement-matched PCA (0.006–0.017); PCA rank sweeps likewise show increasing losses. STS effects remain small. These patterns suggest that transfer costs are associated with intervention extent and geometry. Thus, in-distribution gains do not establish semantic preservation on unseen tasks. For embeddings intended for reuse across tasks or corpora, evaluation should include applications beyond the corpus that motivated the intervention.

\section{Conclusion}

We investigated when document embeddings can be deconfounded without metadata labels while preserving semantic utility. Across six corpora and multiple embedding models, controlled analyzes of geometric and SAE-based erasure reveal the conditions and mechanisms underlying successful label-free deconfounding.


Our findings suggest that effective deconfounding is not simply a matter of discovering confounder-related features, but also of understanding how these features interact with embedding geometry and the interventions used to suppress them. The competitive performance of simple geometric approaches demonstrates that sophisticated feature discovery is not necessarily required to recover substantial semantic benefits. Moreover, poor proxy–confounder alignment and insufficient intervention extent represent distinct obstacles to effective erasure, highlighting the importance of separating feature discovery from intervention design when comparing methods. More fundamentally, confounder suppression and semantic preservation are distinct objectives: reduced linear recoverability does not guaranty the elimination of nonlinear confounder information, while stronger suppression need not improve semantic utility and may compromise performance on unseen tasks.


Together, these findings motivate a more nuanced evaluation of label-free deconfounding that accounts for intervention effects, residual confounder information, and semantic preservation beyond the fitting corpus. However, our post-hoc diagnostics still require confounder labels, leaving reliable label-free selection of methods and intervention strengths unresolved. Future work should investigate how embedding-model architecture, training, and corpus composition influence the separability of nuisance and semantic information, develop label-free criteria for anticipating erasure effectiveness and semantic costs, and establish when residual nonlinear confounder information affects downstream behavior. Ultimately, successful deconfounding should be understood not merely in terms of the information removed, but also in terms of the semantic information preserved.


\section{Limitations}

Our study examines only language and document-source confounders; whether the findings generalize to attributes such as genre, writing style, or author identity remains unexplored. We also do not investigate how possible correlations between confounders and task-relevant semantic information affect erasure effectiveness and utility. Although we assess leakage primarily through linear-probe AUC, substantial nonlinear confounder information remains recoverable even after metadata-supervised LEACE. We do not address its removal, as our analysis is restricted to affine interventions on frozen embeddings, excluding nonlinear erasure and joint adaptation with downstream tasks. Finally, retrieval, semantic similarity, and topic clustering cover only a subset of downstream applications and may overlook losses in other semantic properties.


\clearpage
\bibliography{tacl2021}
\bibliographystyle{acl_natbib}
\clearpage

\FloatBarrier
\appendix

\input{appendix_4/Appendix_A}
\clearpage
\input{appendix_4/Appendix_B}

\end{document}

%% file: tables_4/table_1_datasets.tex
\begin{table*}[t]
\centering
\footnotesize
\renewcommand{\arraystretch}{0.92}
\setlength{\tabcolsep}{4pt}
\setlength{\abovecaptionskip}{2pt}
\setlength{\belowcaptionskip}{0pt}
\begin{tabular*}{\textwidth}{@{\extracolsep{\fill}}llrrrrrl}
\toprule
 &  &  &  & \multicolumn{2}{c}{Linked groups} &  &  \\
\cmidrule(lr){5-6}
Corpus & Nuisance (classes) & C & Docs & pairs & triples & Ret.\ pairs & Semantic evaluation \\
\midrule
SemEval & Language (DE/EN/ES/ZH) & 4 & 856 & 428 & 0 & 428 & Retrieval, graded sim. \\
SwiLTra & Language (DE/FR/IT) & 3 & 8,352 & 0 & 2,784 & 8,352 & Retrieval \\
Fedlex & Language (DE/FR/IT) & 3 & 8,352 & 0 & 2,784 & 8,352 & Retrieval \\
Bills--Tweets & Source (bills/tweets) & 2 & 9,933 & -- & -- & -- & None (leakage only) \\
SCOTUS & Source (Oyez/Lexis/Wiki) & 3 & 7,506 & 1,286 & 633 & 3,185 & Retrieval \\
CAP & Source (bills/orders/news) & 3 & 3,208 & -- & -- & -- & 21-topic clustering \\
\bottomrule
\end{tabular*}
\caption{Datasets and evaluation settings. $C$: number of confounder classes. \emph{Linked groups}: pairs or triples of documents describing the same item, kept intact across training and evaluation. \emph{Ret.\ pairs}: unordered document pairs within linked groups (three per triple), evaluated bidirectionally when held out. Construction details and binary SwiLTra/SCOTUS compositions appear in Appendix~\ref{sec:appendix_A1}. Sources: SemEval \citep{chen2022semeval}; SwiLTra and Fedlex \citep{niklaus2025swiltra}; SCOTUS \citep{fang2023superscotus}; CAP \citep{jones2023codebook}.}
\label{tab:datasets}
\end{table*}

%% file: tables_4/table_2_primary_id.tex
\begin{table*}[t]
\centering
\footnotesize
\renewcommand{\arraystretch}{0.92}
\setlength{\tabcolsep}{3pt}
\setlength{\abovecaptionskip}{2pt}
\setlength{\belowcaptionskip}{0pt}
\begin{tabular*}{\textwidth}{@{\extracolsep{\fill}}lrrrrrr@{\hskip 10pt}r}
\toprule
 & \multicolumn{6}{c}{E5-small} & MiniLM \\
\cmidrule(lr){2-7}\cmidrule(l{6pt}){8-8}
 & SemEval & SwiLTra & Fedlex & Bills--Tweets & SCOTUS & CAP & SCOTUS \\
 & C=4 & C=3 & C=3 & C=2 & C=3 & C=3 & C=3 \\
\midrule
\multicolumn{8}{l}{\textit{Panel A: linear-probe nuisance ROC-AUC $\downarrow$ (one-vs-rest macro average for $C>2$)}} \\
Raw & 1.000{\scriptsize$\,\pm$.000} & .999{\scriptsize$\,\pm$.000} & 1.000{\scriptsize$\,\pm$.000} & 1.000{\scriptsize$\,\pm$.000} & .999{\scriptsize$\,\pm$.000} & 1.000{\scriptsize$\,\pm$.000} & .986{\scriptsize$\,\pm$.000} \\
PCA-1 & .927{\scriptsize$\,\pm$.006} & .906{\scriptsize$\,\pm$.004} & .911{\scriptsize$\,\pm$.006} & .678{\scriptsize$\,\pm$.014} & .882{\scriptsize$\,\pm$.004} & .892{\scriptsize$\,\pm$.008} & .986{\scriptsize$\,\pm$.000} \\
$k$-means LEACE & .797{\scriptsize$\,\pm$.050} & \textbf{.577{\scriptsize$\,\pm$.011}} & \textbf{.644{\scriptsize$\,\pm$.016}} & .612{\scriptsize$\,\pm$.011} & .699{\scriptsize$\,\pm$.085} & .879{\scriptsize$\,\pm$.012} & .950{\scriptsize$\,\pm$.047} \\
Metadata-LEACE & \textbf{.649{\scriptsize$\,\pm$.022}} & \textbf{.577{\scriptsize$\,\pm$.011}} & \textbf{.644{\scriptsize$\,\pm$.016}} & \textbf{.609{\scriptsize$\,\pm$.011}} & \textbf{.615{\scriptsize$\,\pm$.004}} & \textbf{.658{\scriptsize$\,\pm$.024}} & \textbf{.611{\scriptsize$\,\pm$.005}} \\
SAE-LEACE ($N{=}32$) & .699{\scriptsize$\,\pm$.017} & .615{\scriptsize$\,\pm$.010} & .672{\scriptsize$\,\pm$.016} & .653{\scriptsize$\,\pm$.012} & .727{\scriptsize$\,\pm$.009} & .713{\scriptsize$\,\pm$.022} & .860{\scriptsize$\,\pm$.017} \\
\midrule
\multicolumn{8}{l}{\textit{Panel B: semantic utility $\uparrow$ (R@10; CAP: topic purity)}} \\
Raw & .597{\scriptsize$\,\pm$.000} & .492{\scriptsize$\,\pm$.004} & .784{\scriptsize$\,\pm$.009} & \multicolumn{1}{c}{---} & .626{\scriptsize$\,\pm$.005} & .242{\scriptsize$\,\pm$.009} & .697{\scriptsize$\,\pm$.006} \\
PCA-1 & .811{\scriptsize$\,\pm$.003} & .897{\scriptsize$\,\pm$.003} & .950{\scriptsize$\,\pm$.004} & \multicolumn{1}{c}{---} & .721{\scriptsize$\,\pm$.008} & .318{\scriptsize$\,\pm$.019} & .696{\scriptsize$\,\pm$.006} \\
$k$-means LEACE & .814{\scriptsize$\,\pm$.020} & \textbf{.969{\scriptsize$\,\pm$.003}} & \textbf{.981{\scriptsize$\,\pm$.002}} & \multicolumn{1}{c}{---} & .787{\scriptsize$\,\pm$.034} & .320{\scriptsize$\,\pm$.009} & .715{\scriptsize$\,\pm$.023} \\
Metadata-LEACE & \textbf{.877{\scriptsize$\,\pm$.001}} & \textbf{.969{\scriptsize$\,\pm$.003}} & \textbf{.981{\scriptsize$\,\pm$.002}} & \multicolumn{1}{c}{---} & \textbf{.800{\scriptsize$\,\pm$.008}} & \textbf{.352{\scriptsize$\,\pm$.012}} & \textbf{.749{\scriptsize$\,\pm$.006}} \\
SAE-LEACE ($N{=}32$) & .811{\scriptsize$\,\pm$.007} & .916{\scriptsize$\,\pm$.003} & .938{\scriptsize$\,\pm$.005} & \multicolumn{1}{c}{---} & .770{\scriptsize$\,\pm$.005} & .263{\scriptsize$\,\pm$.033} & .712{\scriptsize$\,\pm$.010} \\
\bottomrule
\end{tabular*}
\caption{\textbf{Primary in-distribution leakage and semantic utility.}
Cells report mean $\pm$ sample SD across fitting seeds 0--4 in all settings. SemEval's five outer folds are aggregated within each seed: fold AUCs are averaged and held-out retrieval hits pooled. Only Metadata-LEACE uses nuisance labels for fitting.}
\label{tab:primary-id}
\end{table*}

%% file: tables_4/table3_mechanism_controls.tex
\begin{table*}[t]
\centering
\footnotesize
\setlength{\tabcolsep}{2pt}
\begin{tabular*}{\textwidth}{@{\extracolsep{\fill}}llrrrrrr}
\toprule
 &  & \multicolumn{3}{c}{SwiLTra (DE/FR/IT)} & \multicolumn{3}{c}{SCOTUS (3 sources)} \\
\cmidrule(lr){3-5}\cmidrule(lr){6-8}
 & Condition & Lin.\ AUC$\downarrow$ & MLP AUC$\downarrow$ & R@10$\uparrow$ & Lin.\ AUC$\downarrow$ & MLP AUC$\downarrow$ & R@10$\uparrow$ \\
\midrule 
\multicolumn{8}{l}{\textbf{A. Direction selection (Directions-LEACE; $N{=}32$ except A4)}} \\
A1 & MI-selected SAE directions & \textbf{.616{\scriptsize$\,\pm$.010}} & .838{\scriptsize$\,\pm$.007} & .914{\scriptsize$\,\pm$.004} & \textbf{.727{\scriptsize$\,\pm$.013}} & \textbf{.797{\scriptsize$\,\pm$.014}} & \textbf{.774{\scriptsize$\,\pm$.005}} \\
A2 & Random SAE directions & .630{\scriptsize$\,\pm$.010} & .842{\scriptsize$\,\pm$.007} & .909{\scriptsize$\,\pm$.003} & .742{\scriptsize$\,\pm$.002} & .814{\scriptsize$\,\pm$.005} & .772{\scriptsize$\,\pm$.005} \\
A3 & Random non-SAE directions & .819{\scriptsize$\,\pm$.015} & .913{\scriptsize$\,\pm$.007} & \textbf{.923{\scriptsize$\,\pm$.003}} & .887{\scriptsize$\,\pm$.010} & .916{\scriptsize$\,\pm$.006} & .771{\scriptsize$\,\pm$.006} \\
A4 & Non-SAE directions, displ.-matched & .781{\scriptsize$\,\pm$.012} & .893{\scriptsize$\,\pm$.007} & .911{\scriptsize$\,\pm$.003} & .861{\scriptsize$\,\pm$.009} & .894{\scriptsize$\,\pm$.006} & .767{\scriptsize$\,\pm$.004} \\
\addlinespace[1pt]
A1$^{\dagger}$ & MI-selected, fits of A5--A9 & .617{\scriptsize$\,\pm$.012} & .842{\scriptsize$\,\pm$.009} & .918{\scriptsize$\,\pm$.004} & .737{\scriptsize$\,\pm$.008} & .811{\scriptsize$\,\pm$.006} & .772{\scriptsize$\,\pm$.003} \\
A5$^{\dagger}$ & Entropy-selected SAE directions & .618{\scriptsize$\,\pm$.013} & \textbf{.832{\scriptsize$\,\pm$.015}} & .912{\scriptsize$\,\pm$.006} & .752{\scriptsize$\,\pm$.006} & .819{\scriptsize$\,\pm$.008} & .768{\scriptsize$\,\pm$.003} \\
A6$^{\dagger}$ & Dip-selected SAE directions & .617{\scriptsize$\,\pm$.011} & .840{\scriptsize$\,\pm$.010} & .917{\scriptsize$\,\pm$.005} & .756{\scriptsize$\,\pm$.011} & .826{\scriptsize$\,\pm$.012} & .769{\scriptsize$\,\pm$.004} \\
A7$^{\dagger}$ & Co-firing-selected SAE directions & .623{\scriptsize$\,\pm$.017} & .835{\scriptsize$\,\pm$.006} & .914{\scriptsize$\,\pm$.003} & .751{\scriptsize$\,\pm$.014} & .815{\scriptsize$\,\pm$.007} & .769{\scriptsize$\,\pm$.004} \\
A8$^{\dagger}$ & Co-firing-selected, exclusive pairs & .620{\scriptsize$\,\pm$.014} & .840{\scriptsize$\,\pm$.006} & .917{\scriptsize$\,\pm$.005} & .745{\scriptsize$\,\pm$.009} & .811{\scriptsize$\,\pm$.008} & .769{\scriptsize$\,\pm$.004} \\
A9$^{\dagger}$ & Oracle (MI with the true classes) & \textbf{.616{\scriptsize$\,\pm$.011}} & .841{\scriptsize$\,\pm$.009} & .918{\scriptsize$\,\pm$.004} & .732{\scriptsize$\,\pm$.003} & .806{\scriptsize$\,\pm$.005} & .773{\scriptsize$\,\pm$.004} \\
\addlinespace[1pt]
\midrule
\multicolumn{8}{l}{\textbf{B. Operator (same MI-selected latent set $S$ as row A1)}} \\
B1 & SAE activations + Activation-LEACE & .812{\scriptsize$\,\pm$.018} & .901{\scriptsize$\,\pm$.009} & \textbf{.933{\scriptsize$\,\pm$.004}} & .862{\scriptsize$\,\pm$.005} & .896{\scriptsize$\,\pm$.006} & .768{\scriptsize$\,\pm$.003} \\
B2 & SAE directions + orthogonal removal & .998{\scriptsize$\,\pm$.001} & .999{\scriptsize$\,\pm$.000} & .894{\scriptsize$\,\pm$.017} & .995{\scriptsize$\,\pm$.001} & .996{\scriptsize$\,\pm$.001} & .746{\scriptsize$\,\pm$.008} \\
B3 & Orthogonal removal, mean kept & .998{\scriptsize$\,\pm$.001} & .999{\scriptsize$\,\pm$.000} & .894{\scriptsize$\,\pm$.017} & .995{\scriptsize$\,\pm$.001} & .996{\scriptsize$\,\pm$.001} & .746{\scriptsize$\,\pm$.008} \\
B4 & Orth.\ removal of $\mathrm{span}(\Sigma D_S^{\top})$ & \textbf{.616{\scriptsize$\,\pm$.010}} & \textbf{.838{\scriptsize$\,\pm$.006}} & .914{\scriptsize$\,\pm$.004} & \textbf{.727{\scriptsize$\,\pm$.013}} & \textbf{.796{\scriptsize$\,\pm$.011}} & \textbf{.774{\scriptsize$\,\pm$.005}} \\
\addlinespace[1pt]
\midrule
\multicolumn{8}{l}{\textbf{C. Erasure extent}} \\
C1 & PCA, displacement-matched & .600{\scriptsize$\,\pm$.013} & .817{\scriptsize$\,\pm$.010} & .882{\scriptsize$\,\pm$.005} & .677{\scriptsize$\,\pm$.003} & .756{\scriptsize$\,\pm$.006} & .774{\scriptsize$\,\pm$.005} \\
\addlinespace[1pt]
\midrule
\multicolumn{8}{l}{\textbf{References}} \\[1pt]
 & Raw & .999{\scriptsize$\,\pm$.000} & .999{\scriptsize$\,\pm$.000} & .492{\scriptsize$\,\pm$.004} & .999{\scriptsize$\,\pm$.000} & .999{\scriptsize$\,\pm$.000} & .626{\scriptsize$\,\pm$.005} \\
 & Metadata-LEACE (supervised) & .577{\scriptsize$\,\pm$.011} & .923{\scriptsize$\,\pm$.003} & .969{\scriptsize$\,\pm$.003} & .615{\scriptsize$\,\pm$.004} & .880{\scriptsize$\,\pm$.008} & .800{\scriptsize$\,\pm$.008} \\
\bottomrule
\end{tabular*}
\caption{\textbf{Direction selection, erasure operators, and extent.}
E5-small results on three-class SwiLTra and SCOTUS (mean $\pm$ sample SD, seeds 0--4). AUCs are held-out macro one-vs-rest; R@10 measures paired retrieval. Row definitions and matching procedures are detailed in Appendices~\ref{sec:appendix_selection}--\ref{sec:appendix_extent}; random controls average five draws per seed. $\dagger$ rows use separate SAE fits, with A1$^\dagger$ as their matched baseline. Reference rows reproduce Table~\ref{tab:primary-id}.}
\label{tab:mechanism-controls}
\end{table*}

%% file: tables_4/table_erasure_extent.tex
\begin{table*}[!t]
\centering
\scriptsize
\renewcommand{\arraystretch}{0.88}
\setlength{\tabcolsep}{1.5pt}
\setlength{\abovecaptionskip}{1pt}
\setlength{\belowcaptionskip}{0pt}
\resizebox{0.92\textwidth}{!}{%
\begin{tabular}{lccrrlrrl}
\toprule
 &  &  & \multicolumn{3}{c}{Linear AUC$\downarrow$} & \multicolumn{3}{c}{Utility$\uparrow$} \\
\cmidrule(lr){4-6}\cmidrule(lr){7-9}
Setting & Displ. & $r^{*}$ & SAE & PCA & PCA $-$ SAE & SAE & PCA & PCA $-$ SAE \\
\midrule
SwiLTra (DE/FR/IT) & .36 & 22--23 & .616 & \textbf{.600} & $-$.016{\scriptsize\,[$-$.021, $-$.011]} & \textbf{.914} & .882 & $-$.032{\scriptsize\,[$-$.037, $-$.026]} \\
SCOTUS (3 sources) & .30 & 22 & .727 & \textbf{.677} & $-$.049{\scriptsize\,[$-$.065, $-$.033]} & \textbf{.774} & \textbf{.774} & +.001{\scriptsize\,[$-$.003, +.005]} \\
SemEval & .42 & 18--21 & .702 & \textbf{.684} & $-$.018{\scriptsize\,[$-$.035, .000]} & .807 & \textbf{.814} & +.007{\scriptsize\,[$-$.001, +.015]} \\
CAP & .33 & 20--21 & .713 & \textbf{.687} & $-$.026{\scriptsize\,[$-$.035, $-$.017]} & \textbf{.284} & .215 & $-$.069{\scriptsize\,[$-$.135, $-$.003]} \\
Fedlex (DE/FR/IT) & .36 & 21 & .674 & \textbf{.665} & $-$.009{\scriptsize\,[$-$.015, $-$.002]} & \textbf{.939} & .926 & $-$.013{\scriptsize\,[$-$.019, $-$.008]} \\
SCOTUS (3 sources), MiniLM & .36 & 19--21 & .840 & \textbf{.789} & $-$.051{\scriptsize\,[$-$.078, $-$.023]} & .713 & \textbf{.735} & +.022{\scriptsize\,[+.012, +.031]} \\
\bottomrule
\end{tabular}%
}
\caption{\textbf{Erasure extent: SAE-LEACE vs.\ displacement-matched PCA.} E5-small unless marked; mean results over seeds 0--4. PCA uses the smallest rank $r^{*}$ whose normalized squared training displacement reaches that of SAE-LEACE ($N=32$). PCA $-$ SAE reports the paired per-seed difference with 95\% $t$-intervals. Utility is R@10, except CAP topic purity. The first two rows correspond to A1/C1 in Table~\ref{tab:mechanism-controls}.}
\label{tab:erasure-extent}
\end{table*}

%% file: appendix_4/Appendix_A.tex
\section{Methodology Appendix}
\label{sec:appendix_A}

\subsection{Dataset Construction and Composition}
\label{sec:appendix_A1}

\input{tables_4/tableA1_dataset_composition}

Table~\ref{tab:dataset-composition} gives the processed SwiLTra, Fedlex, and SCOTUS compositions before splitting.

\paragraph{SwiLTra and SCOTUS.} In SwiLTra \citep{niklaus2025swiltra}, each aligned triple contains the same Swiss Federal Supreme Court decision headnote in German, French, and Italian, yielding three semantically equivalent versions that differ by language.

SCOTUS combines U.S. Supreme Court case summaries from LexisNexis and Oyez \citep{fang2023superscotus} with Wikipedia summaries from \citet{fan-etal-2025-medium}. Here, each complete triple contains three summaries of the same case, one from each source. We pair corresponding rows within the first 1,024 positions of each source block, except for Oyez–Wikipedia, for which the first 780 are used. We remove exact duplicate texts, discard links missing a member, and retain the remaining documents as distractors. The three-source setting pools the pairwise files, merges identical $(\text{source},\text{text})$ documents, and joins compatible links transitively. Components containing multiple summaries from the same source are left unpaired; thus, ``unpaired'' denotes the absence of a retained link, not a verified absence of a counterpart.

\paragraph{Fedlex.} We use the article-level CH-Law-Trans data from SwiLTra-Bench \citep{niklaus2025swiltra}, derived from Swiss federal legislation published on Fedlex. Each aligned sample contains the same Swiss legal article in German, French, and Italian. We retain domestic-law articles with non-empty versions in all three languages, excluding international law (SR~0.*) and cantonal constitutions (SR~131.2). Processing records in file order, we discard any aligned triple containing two identical texts or a text already retained in any language. This removes 4,654 of 71,564 complete triples, leaving 66,910 intact triples with globally unique texts. We then sample 2,784 articles (seed 42), yielding 8,352 documents and matching trilingual SwiLTra in both size and language composition.

\paragraph{CAP.} We construct the CAP corpus from the Comparative Agendas Project \citep{jones2023codebook}. The corpus comprises documents from three sources—congressional bill summaries, executive orders, and newspaper articles—and each document is assigned one of twenty-one major topic labels. Within each source, we remove exact duplicates, shuffle the documents (seed 42), and retain at most 64 documents per topic. Executive orders are then truncated to the first 598 sampled documents, following the count implied by Table~1 of \citet{fan-etal-2025-medium}. The resulting 3,208 documents comprise 1,301 congressional bill summaries \citep{wilkerson2023bills,hoyle-etal-2022-neural}, 598 executive orders \citep{jones2023orders}, and 1,309 newspaper articles.\footnote{https://comparativeagendas.net/project/pennsylvania} Executive orders span 20 of the 21 topics, with no documents labeled as Culture. Topic labels are used only for sampling and semantic-utility evaluation and never for eraser fitting.

\paragraph{SemEval.} We use the 444 English–non--English news-article pairs from \citet{chen2022semeval}. Each pair consists of two independently written articles describing the same underlying event, rather than translations of one another, with article bodies taken from \texttt{text1} and \texttt{text2}. After removing 16 pairs that reuse an article, 428 pairs remain: 173 German–English, 189 Spanish–English, and 66 Chinese–English. Human similarity ratings range from 1 (very similar) to 4 (very dissimilar); for graded similarity evaluation, we use $-\texttt{Overall}$ without further rescaling.

\paragraph{Bills--Tweets.} We use the Bills–Tweets dataset distributed with the repository of \citet{fan-etal-2025-medium}, although it was not analyzed in their paper. It comprises documents from two sources: 5,000 Congressional Research Service bill summaries from the 110th–114th Congresses \citep{hoyle-etal-2022-neural} and 5,000 original tweets from the Tweets of Congress archive, posted by members of the 115th Congress through official or campaign accounts between June 2017 and January 2019. Removing 65 duplicate bill summaries and two duplicate tweets leaves 9,933 documents; no further filtering or balancing is applied. The dataset provides source labels but no semantic-utility annotations, so it is used only for confounder-leakage evaluation.

\subsection{Models Configurations}
\label{sec:appendix_A2}

\input{tables_4/tableA2_encoders}

Table~\ref{tab:encoders} specifies models, prompts, and sequence lengths. Methods share embedding model outputs within each setting, with maps fitted separately for each model. The SAE adds no input normalization; transformed outputs are not renormalized. All similarity scoring uses cosine similarity.


The refitting study uses all ten embedding models across all datasets: SemEval, binary (DE–FR), and trilingual SwiLTra, Bills–Tweets, binary (Oyez–Lexis), and three-source SCOTUS, CAP, and Fedlex.

\subsection{Geometric Erasure and Supervised Reference}
\label{sec:appendix_geometric}

Table~\ref{tab:hyperparameters} supplies numerical settings for the methods in Section~\ref{sec:baselines}, SAE training, controls, and probes.

\begin{table*}[t]
\centering
\footnotesize
\setlength{\tabcolsep}{3pt}
\renewcommand{\arraystretch}{1.02}
\begin{tabular}{@{}>{\raggedright\arraybackslash}p{0.145\textwidth} >{\raggedright\arraybackslash}p{0.825\textwidth}@{}}
\toprule
Component & Setting \\
\midrule
\textbf{SAE training} & Adam; learning rate $10^{-3}$; batch size 1,024; 200 epochs. Initially $W_{\mathrm{enc}}=W_{\mathrm{dec}}^\top$ and $b_{\mathrm{pre}}=\mu$; decoder rows have unit norm after each step. AuxK: $k_{\mathrm{aux}}=2k$, $\alpha_{\mathrm{aux}}=1/32$; a latent is dead after 20 epochs without firing. \\
\addlinespace[0.15em]
\textbf{Pseudo-groups} & Scikit-learn $k$-means, $K\in\{2,3,4,5,6,8\}$. Maximize silhouette on at most 4,000 training documents; ties favor smaller $K$. \\
\addlinespace[0.15em]
\textbf{LEACE} & Continuous concept columns are standardized using training statistics. \\
\addlinespace[0.15em]
\textbf{Orthogonal bases} & Uncentered decoder-span removal: SVD cut-off $10^{-6}$. Mean-preserving variants: basis cut-off $10^{-8}$ times the largest singular value. \\
\addlinespace[0.15em]
\textbf{Sweeps} & Figure~\ref{fig:fig_1}: $r\in\{1,2,3,4,8,16,32,64\}$, $K$ as above, $N\in\{1,2,4,8,12,16,24,32,48,64\}$. Frozen-transfer PCA sweep: the same $r$ grid without 64. \\
\addlinespace[0.15em]
\textbf{Probes} & Logistic regression: at most 5,000 iterations; regularization $C\in\{0.01,0.1,1,10,100\}$. Standardized features; coordinates are dropped if their SD is below $10^{-6}$ times the median SD. MLP: hidden layers $(256,128)$, at most 300 iterations, early stopping. \\
\addlinespace[0.15em]
\textbf{Utility} & CAP topic $k$-means uses 20 initializations. \\
\bottomrule
\end{tabular}
\caption{Numerical settings. Primary SAE dimensions and budget are defined in Section~\ref{sec:sae-leace}; fitting and probe nesting in Appendix~\ref{sec:appendix_id}.}
\label{tab:hyperparameters}
\end{table*}

\subsection{SAE-Based Erasure}
\label{sec:appendix_A3}

\paragraph{Training.}
For row-vector input $x$, the SAE computes
\begin{equation}
\begin{aligned}
h &= (x-b_{\mathrm{pre}})W_{\mathrm{enc}}+b_{\mathrm{enc}},\\
z &= \operatorname{ReLU}(\operatorname{TopK}_k(h)),\\
\hat{x} &= zW_{\mathrm{dec}}+b_{\mathrm{pre}}.
\end{aligned}
\end{equation}
AuxK \citep{gao2025scaling} reconstructs the residual $e=x-\hat{x}$ as $\hat e_{\mathrm{aux}}$ using the highest-activating dead latents. The per-example loss,
\begin{equation}
\mathcal L_{\mathrm{SAE}}=\frac{1}{d}\left(
\|x-\hat{x}\|_2^2+
\alpha_{\mathrm{aux}}\|e-\hat e_{\mathrm{aux}}\|_2^2\right),
\end{equation}
is averaged over each minibatch; optimization, auxiliary budget, and inactivity threshold are specified in Table~\ref{tab:hyperparameters}.

\paragraph{Firing frequencies and MI ranking.}
For $a_{ij}=\mathbf{1}[F_{\mathrm{tr},ij}>0]$ and pseudo-group $g$, training frequencies are defined as follows:
\begin{equation}
\widehat I_j=\sum_{a\in\{0,1\}}\sum_g
\widehat p_j(a,g)\log_2
\frac{\widehat p_j(a,g)}{\widehat p_j(a)\widehat p(g)},
\end{equation}
omitting zero-probability terms. Latents outside the training firing-rate band $[0.02,0.98]$ score zero rather than being excluded. Direct index checks confirm that all top 64 latents lie inside the band in the 20 fits of Figure~\ref{fig:fig_1} (64–200 eligible latents per fit), as do all top 32 in the reported comparisons of Tables~\ref{tab:mechanism-controls}, \ref{tab:erasure-extent}, and~\ref{tab:sae-construction}. Thus, MI and random controls select from the same pool at every reported budget. Scored rankings descend by score, breaking ties by lower latent index; MI sweeps use nested prefixes.

\subsection{Controlled Comparisons}
\label{sec:appendix_A4}

Table~\ref{tab:mechanism-controls} uses E5-small on three-class SwiLTra and SCOTUS. Candidate-selection arms use Directions-LEACE; random controls average five draws per fitting seed.

\subsubsection{Candidate-Direction Selection}
\label{sec:appendix_selection}

Throughout this subsection, row labels refer to Table~\ref{tab:mechanism-controls}.

\paragraph{Random SAE directions.} Row~A2 samples uniformly without replacement from the firing-rate-eligible pool defined in Appendix~\ref{sec:appendix_A3}, with nested budgets within each draw.

\paragraph{Rankings without $k$-means.}
Rows~A5--A8 rank the same eligible pool using criteria that do not depend on the $k$-means partition. For training firing rate $\widehat p_j$, decoder row $d_j$, and pool size $M$, firing entropy is
\[
H(a_j)=-\widehat p_j\log_2\widehat p_j-(1-\widehat p_j)\log_2(1-\widehat p_j).
\]
Because $\widehat I_j\leq H(a_j)$, this controls for marginal firing rate. Hartigan's dip statistic \citep{hartigan1985dip}, computed on the training coordinates $X_{\mathrm{tr}}d_j^\top$, measures the departure from unimodality along the decoder direction.

The co-firing criteria rank latents based on how strongly their binary activation patterns are statistically associated with those of other eligible latents. For each latent $j$, we compute the pairwise mutual information between its firing indicator $a_j$ and every other latent $a_l$, and average these associations:
\begin{equation}
c_j=\frac{1}{M-1}\sum_{l\neq j}\widehat I(a_j;a_l).
\end{equation}
A high score therefore identifies latents whose firing behavior shares substantial information with many other latents, providing a purely activation-based measure of coordinated structure. The second co-firing variant is more selective: it includes only pairs whose firing indicators are negatively correlated while retaining the denominator $M-1$, thereby emphasizing latents that participate in complementary rather than positively co-activated firing patterns. In both cases, MI is estimated from training firing indicators only.

The top $N$ latents are selected without label-tuned hyperparameters. Row~A9 instead ranks by $\widehat I(a_j;y)$, where $y$ denotes the true training confounder classes and therefore serves as a supervised oracle.

\paragraph{Random non-SAE directions.} Row~A3 replaces SAE directions with 32 data- and label-independent rows drawn from $\mathcal N(0,I_d)$, scaled to unit norm but not orthogonalized. Their normalized squared training displacement is 0.25--0.32, compared with 0.30--0.36 for SAE-LEACE, so $N$ and the intervention operator are matched, but displacement is not. Row~A4 instead uses the fewest leading random directions whose map reaches or exceeds the displacement of the corresponding MI-selected SAE-LEACE fit (Appendix~\ref{sec:appendix_extent}): 40--43 directions on SwiLTra and 40--41 on SCOTUS.

\subsubsection{Intervention Construction}
\label{sec:appendix_operators}

With the MI-selected $S$ fixed, Activation-LEACE supplies $F_{\mathrm{tr},S}$ instead of the directional coordinates of Section~\ref{sec:sae-leace}. For orthonormal basis rows $Q$, orthogonal removal is
\begin{equation}
\begin{aligned}
T_\perp(x)&=x(I-Q^\top Q),\\
T_{\perp,\mu}(x)&=\mu+(x-\mu)(I-Q^\top Q).
\end{aligned}
\end{equation}
Rows~B2 and~B3 use a basis of $\mathrm{span}(D_S)$, respectively, without and with mean restoration. Row~B4 uses the mean-preserving map with a basis of $\mathrm{span}(\widehat\Sigma D_S^\top)$, where $\widehat\Sigma$ is the training covariance. This is the Directions-LEACE displacement subspace: its largest principal angle to the fitted map's row space of $I-A$ is below $10^{-3}$ degrees in every Table~\ref{tab:mechanism-controls} fit. Basis tolerances follow Table~\ref{tab:hyperparameters}.

\subsubsection{Erasure Extent}
\label{sec:appendix_extent}


\paragraph{Effective rank.} The rank of $I-A$ counts the linearly independent embedding directions modified by the map. For PCA, removing $r$ orthonormal directions yields a rank of $r$ by construction. For LEACE, the rank is bounded by that of the embedding–concept cross-covariance. Centered one-hot concepts, therefore, yield upper bounds of $K-1$ and $C-1$ for $k$-means LEACE and Metadata-LEACE, respectively, while the $N$ continuous SAE concepts yield a bound of $N$.

We verify that these bounds are attained across 900 fitted maps from seeds 0–2 of Figure~\ref{fig:fig_1}. Each of the four settings and three seeds contributes 8 PCA ranks, 6 cluster counts, 10 MI budgets, 10 random-SAE budgets with five draws each, and Metadata-LEACE. The numerical rank of $I-A$ equals $r$, $K-1$, $C-1$, and $N$ for the respective methods: nonzero singular values are at least 1 (up to rounding), while all others are below $10^{-6}$. Saved $N=32$ SAE-LEACE maps for all six Figure~\ref{fig:fig_2} settings also have rank 32.

\paragraph{Displacement matching.}
For $n$ training, embeddings centered in $X_c$ and the fitted $N=32$ MI-selected SAE-LEACE map,
\begin{equation}
\begin{aligned}
\widehat\Sigma &= X_c^\top X_c/n,\\
\nu_{\mathrm{SAE}} &=
\frac{\operatorname{tr}[(I-A)^\top\widehat\Sigma(I-A)]}
{\operatorname{tr}(\widehat\Sigma)}.
\end{aligned}
\end{equation}
When $b=\mu(I-A)$, this equals
\[
\nu_{\mathrm{SAE}}=
\frac{\sum_i\|x_i-T(x_i)\|^2}{\sum_i\|x_i-\mu\|^2}.
\]
The fitted three-class maps preserve the training mean to numerical precision. We use float64-centered training embeddings and the final map after LEACE's numerical constraints, without additional normalization or clipping.

For descending training-covariance eigenvalues $\lambda_j$, choose
\begin{equation}
\begin{aligned}
F(r)&=\frac{\sum_{j=1}^r\lambda_j}{\sum_j\lambda_j},\\
r^\star&=\min\{r:F(r)\geq\nu_{\mathrm{SAE}}-10^{-6}\}.
\end{aligned}
\end{equation}
We search all available PCA ranks separately for each training split, including each SemEval outer fold, and apply the PCA map of Section~\ref{sec:baselines}. For PCA, displacement equals the variance fraction removed; this need not hold for oblique SAE-LEACE. Table~\ref{tab:erasure-extent} reports the six-setting matched comparison. The same displacement statistic governs random-direction matching, frozen-transfer PCA matching, and the SAE-construction control.

\subsubsection{SAE Construction}
\label{sec:appendix_construction}

Table~\ref{tab:sae-construction} repeats MI-selected, random-SAE, and displacement-matched PCA comparisons at $N=32$ on the same two corpora and embedding model, using $(m,k)\in\{(4d,16),(16d,64),(8d,64)\}$. Other training settings, including $k_{\mathrm{aux}}=2k$ and $\alpha_{\mathrm{aux}}$, are unchanged. The table reports eligible-pool sizes; SCOTUS with $(4d,16)$ is not compared because seed 0 leaves only 22 eligible latents, below the required 32.

\subsection{In-Distribution Evaluation}
\label{sec:appendix_id}

\paragraph{Outer splits.}
\label{par:outer-splits}
SemEval uses five outer folds; other corpora use repeated 50/50 splits. Linked groups remain intact; unpaired documents are split individually with confounder-class stratification. Fedlex groups by whole law, keeping every article and aligned triple from a law in one partition.

\paragraph{Seeds and provenance.}
Unless specified otherwise, fitting uses seeds 0–4. The study of the ten embedding models, SAE-construction control, and rank diagnostic uses 0–2. All within-fit randomness (splits, initialization, $k$-means, random selections/directions, and probes) follows the fitting seed. Within runs, matched conditions share splits and probe folds; SAE-based conditions also share the fitted dictionary.

SAE refits are deterministic within a fixed device/software environment. Seeds fix splits and initialization, but minibatch order is drawn on the training device, and floating-point results differ across devices. Primary fits (Table~\ref{tab:primary-id}) and control fits (Tables~\ref{tab:mechanism-controls} and~\ref{tab:erasure-extent}) use two separate GPU runtimes; the $\dagger$ rows use CPU fits. Across these blocks, equal seeds identify the same data splits and probe folds, not identical dictionaries. Row~A1$^\dagger$ recomputes MI ranking on the fits of A5--A9 and is their matched baseline.

\paragraph{Leakage audits.}
\label{par:leakage-audits}
Fresh probes use stratified group five-fold cross-validation within transformed outer-held out documents. Class coverage requires four folds in four of SemEval's five seeds. For Fedlex, probe folds and the inner regularization split preserve article triples, not whole laws: probes may train and test on different articles of the same law, although the eraser was fitted on other laws. Logistic-regression regularization is selected on an inner 80/20 split preserving the same groups. Scaling and near-constant coordinate handling use probe-training statistics; numerical settings appear in Table~\ref{tab:hyperparameters}.

\paragraph{Metadata-LEACE sanity check.}
With centered outer-training embeddings $X_c$ and centered one-hot labels $Y_c$, we measure $\widehat\Sigma_{XY}=X_c^\top Y_c/n$ before and after erasure, reporting
\[
\|\widehat\Sigma_{XY}^{\mathrm{erased}}\|_F/
\|\widehat\Sigma_{XY}^{\mathrm{raw}}\|_F.
\]
Across the 55 primary fits of Table~\ref{tab:primary-id} (one per seed; SemEval: one per outer fold), setting-wise median residuals are $1.5$--$4.0\times10^{-6}$, with maximum $4.1\times10^{-6}$. Repeating one fit per setting in float64 gives $3.0$--$7.9\times10^{-15}$, identifying the float32 residual as numerical rounding rather than failure to erase the training moments.

\paragraph{Utility.}
Each held-out linked pair is evaluated bidirectionally, excluding the query and other known counterparts when ranking its designated target. Unpaired documents are distractors, never targets. R@1/10 averages directional pair-level outcomes, not equally weighted language/source pair types. CAP clusters transformed held-out documents with $K_{\mathrm{topic}}=21$; initializations and retrieval tie handling follow Table~\ref{tab:hyperparameters}.

\paragraph{Aggregation and uncertainty.}
Probe-fold AUCs, and SemEval outer-fold AUCs, are averaged within each seed. SemEval retrieval hits and rated-pair cosines are pooled across outer folds before computing retrieval and Spearman scores. We report mean $\pm$ sample SD across seeds, without pooling their outcomes. Tables~\ref{tab:erasure-extent} and~\ref{tab:sae-construction} form paired differences within each seed, using identical fitting splits and audit documents, and report their mean and 95\% $t$-interval ($n_{\mathrm{seeds}}-1$ degrees of freedom). These intervals describe split-and-fit variability, not within-split document resampling.

\subsection{Post-Hoc Geometry Diagnostics}
\label{sec:appendix_geometry}

Diagnostics use training embeddings and labels from the splits and seeds of their accompanying erasers. For centered embeddings $X_c$ and centered one-hot labels $Y_c$, define
\begin{equation}
\mathcal L=\operatorname{col}(X_c^\top Y_c),
\qquad q=\dim(\mathcal L)\leq C-1.
\end{equation}
This captures linear label covariance, not potentially nonlinear confounder information. With orthonormal basis $Q_{\mathcal L}\in\mathbb R^{d\times q}$ and principal-direction rows $B_r\in\mathbb R^{r\times d}$, PCA capture is $\|B_rQ_{\mathcal L}\|_F^2/q$, equal to 1 when $\mathcal L\subseteq\mathrm{span}(B_r)$.

Cluster--class ARI compares silhouette-selected $k$-means assignments with true confounder classes. Figure~\ref{fig:fig_2} reports PCA capture at $r=C-1$, cluster--class ARI, and each label-free eraser's held-out linear AUC minus Metadata-LEACE's. Table~\ref{tab:ten-encoder-robustness} reports the first two across embedding models for the three-class settings.

\subsection{OOD Targets and Frozen Donor Maps}
\label{sec:appendix_A5}

\paragraph{Targets and donors.}
We evaluate frozen transfer on five MTEB legal retrieval tasks—AILA Casedocs, AILA Statutes, LegalBench Consumer Contracts QA, LegalBench Corporate Lobbying, and Legal Summarization—together with STSBenchmark. We use the MTEB test splits and fixed dataset revisions. Primary donor maps are fitted on CAP and three-source SCOTUS with E5-small and MiniLM, and on trilingual SwiLTra with E5-small, using seeds 0–4. We transfer PCA–1, $k$-means LEACE, Metadata-LEACE, SAE-LEACE, and displacement-matched PCA. A separate PCA-rank sweep uses $r\in\{1,2,3,4,8,16,32\}$ with CAP and binary SCOTUS/SwiLTra donors. No exact case- or whitespace-normalized donor–target text matches were found.

\paragraph{Scoring and uncertainty.}
Frozen maps are applied unchanged to both retrieval queries and documents, and to both STS sentences, without target-specific refitting or output renormalization. Legal retrieval uses cosine nDCG@10, averaged equally across the five tasks; STSBenchmark uses Spearman correlation. We report changes from Raw averaged across fitting seeds. Paired 95\% intervals use 10,000 bootstrap resamples over target queries or STS sentence pairs; maps remain fixed, so these intervals reflect evaluation-sample uncertainty.

\paragraph{Prompt robustness.}
Because E5-small donor inputs use the prefix \texttt{query:\textvisiblespace} whereas OOD targets are unprompted, we repeat the primary E5-small transfer evaluation with the same prefix applied to all target texts (seeds 0--2). The main conclusions are unchanged, indicating that the observed transfer effects are not driven by this prompt mismatch.

%% file: tables_4/tableA1_dataset_composition.tex
\begin{table}[H]
\centering
\small
\setlength{\tabcolsep}{3.5pt}
{\footnotesize%
\begin{tabular}{@{}lrrrrrr@{}}
\toprule
Setting & \multicolumn{3}{c}{Docs per class} & Docs & Pairs & Unpaired \\
\midrule
\textbf{SwiLTra} & \textit{DE} & \textit{FR} & \textit{IT} &  &  &  \\
DE--FR & 2{,}784 & 2{,}784 & -- & 5{,}568 & 2{,}784 & 0 \\
DE--IT & 2{,}784 & -- & 2{,}784 & 5{,}568 & 2{,}784 & 0 \\
FR--IT & -- & 2{,}784 & 2{,}784 & 5{,}568 & 2{,}784 & 0 \\
DE--FR--IT & 2{,}784 & 2{,}784 & 2{,}784 & 8{,}352 & 8{,}352 & 0 \\
\midrule
\textbf{Fedlex} & \textit{DE} & \textit{FR} & \textit{IT} &  &  &  \\
DE--FR--IT & 2{,}784 & 2{,}784 & 2{,}784 & 8{,}352 & 8{,}352 & 0 \\
\midrule
\textbf{SCOTUS} & \textit{Oyez} & \textit{Lexis} & \textit{Wiki} &  &  &  \\
O--L & 2{,}067 & 2{,}034 & -- & 4{,}101 & 1{,}017 & 2{,}067 \\
O--W & 1{,}800 & -- & 1{,}497 & 3{,}297 & 772 & 1{,}753 \\
L--W & -- & 2{,}039 & 1{,}497 & 3{,}536 & 1{,}017 & 1{,}502 \\
O--L--W & 2{,}490 & 3{,}519 & 1{,}497 & 7{,}506 & 3{,}185 & 3{,}035 \\
\bottomrule
\end{tabular}}
\caption{Setting composition after exact-duplicate removal. SCOTUS sources are Oyez (O), LexisNexis (L), and Wikipedia (W); -- denotes an absent class. \emph{Pairs} are linked retrieval pairs; \emph{Unpaired} documents have no recorded link and remain available for fitting and as distractors. Linked groups are kept intact across train/evaluation splits. DE--FR--IT contains 2{,}784 triples in SwiLTra (one per decision) and in Fedlex (one per law article); O--L--W contains 1{,}286 pairs and 633 triples. SCOTUS links follow row alignment as in \citet{fan-etal-2025-medium}; the three-source setting merges identical (source, text) documents and joins links transitively.}
\label{tab:dataset-composition}
\end{table}

%% file: tables_4/tableA2_encoders.tex
\begin{table}[t]
\centering
\small
\setlength{\tabcolsep}{4pt}
{\footnotesize%
\newcolumntype{Y}{>{\raggedright\arraybackslash}X}%
\begin{tabularx}{\columnwidth}{@{}lY r@{}}
\toprule
Encoder & Hugging Face model & Dim. \\
\midrule
MiniLM & {\scriptsize\path{sentence-transformers/all-MiniLM-L6-v2}} & 384 \\
E5-small$^a$ & {\scriptsize\path{intfloat/multilingual-e5-small}} & 384 \\
\midrule
GIST-small & {\scriptsize\path{avsolatorio/GIST-small-Embedding-v0}} & 384 \\
GIST-base & {\scriptsize\path{avsolatorio/GIST-Embedding-v0}} & 768 \\
GIST-large & {\scriptsize\path{avsolatorio/GIST-large-Embedding-v0}} & 1024 \\
E5-base$^a$ & {\scriptsize\path{intfloat/multilingual-e5-base}} & 768 \\
E5-large$^a$ & {\scriptsize\path{intfloat/multilingual-e5-large}} & 1024 \\
MPNet & {\scriptsize\path{sentence-transformers/all-mpnet-base-v2}} & 768 \\
Nomic-v2$^b$ & {\scriptsize\path{nomic-ai/nomic-embed-text-v2-moe}} & 768 \\
MXB & {\scriptsize\path{mixedbread-ai/mxbai-embed-large-v1}} & 1024 \\
\bottomrule
\end{tabularx}}
\caption{Embedding models configurations. Models above the rule are also used in the primary and OOD experiments; those below only in the ten-model study. All are loaded with Sentence-Transformers \citep{reimers-gurevych-2019-sentence}; similarity is cosine. In-distribution inputs use the indicated prompt ($^a$the prefix \texttt{query:\textvisiblespace}; $^b$the named prompt \texttt{passage}); OOD targets are unprompted (a prompt-matched check is in Appendix~\ref{sec:appendix_A5}). Maximum sequence length is each model's default.}
\footnotetext{\url{https://www.sbert.net/}}
\label{tab:encoders}
\end{table}

%% file: appendix_4/Appendix_B.tex
\twocolumn[\mbox{}
\section{Additional results}
\label{sec:appendix_B}
\vspace{0.2em}
\input{tables_4/tableB1_extended_id_block}
\vspace{1em}
]
\clearpage
\onecolumn

\input{tables_4/tableB2_ten_encoder_robustness_singlepage}

\clearpage
\begin{figure}[H]
\centering
\includegraphics[width=0.88\textwidth]{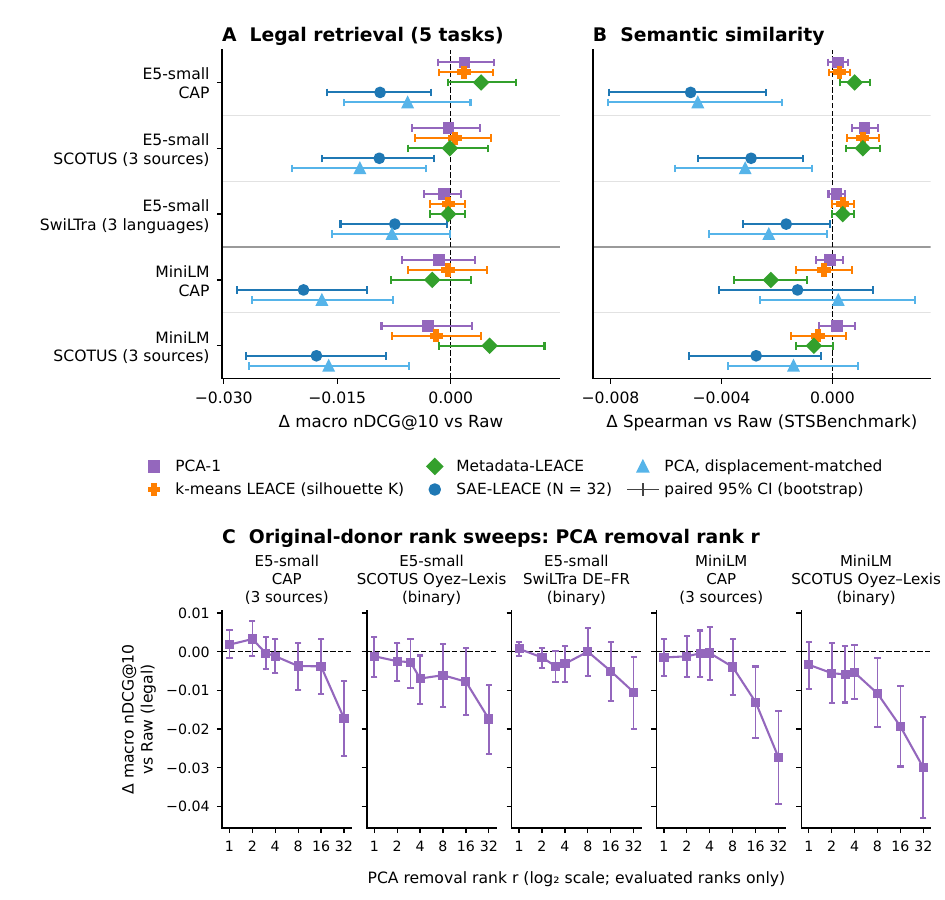}
\captionsetup{skip=2pt}
\caption{\textbf{Frozen transfer reveals rank-dependent utility costs.} Changes from Raw are shown for legal retrieval (macro nDCG@10), STSBenchmark, and PCA rank sweeps under frozen donor maps. Broader interventions generally incur larger transfer losses; SAE–LEACE is comparable to other higher-rank erasers. Points show seed means; bars show paired 95\% bootstrap intervals.}
\label{fig:fig_B3}
\vspace{0.8em}
\end{figure}

\input{tables_4/table_SAE_construction_block}

\twocolumn

\twocolumn

%% file: tables_4/tableB1_extended_id_block.tex
\par\vspace{28pt}\noindent
\begingroup
\centering
\fontsize{9.5pt}{11.3pt}\selectfont
\setlength{\tabcolsep}{3.2pt}%
\renewcommand{\arraystretch}{1.11}%
\newsavebox{\extendedIDsafebox}%
\sbox{\extendedIDsafebox}{%
\begin{tabular}{@{}ll@{\hspace{7pt}}rrrrr@{}}
\toprule
Setting & Metric & Raw & PCA-1 & \shortstack{$k$-means\\LEACE} & \shortstack{Metadata-\\LEACE} & \shortstack{SAE-LEACE\\($N{=}32$)} \\
\midrule
\addlinespace[2pt]
\multicolumn{7}{@{}l}{\textbf{SwiLTra} \quad {\footnotesize confounder: language; 5 seeds (0--4), primary cohort}} \\
\cmidrule(lr){1-7}
\textbf{DE--FR} & Linear AUC $\downarrow$ & .999{\scriptsize$\,\pm$.000} & .632{\scriptsize$\,\pm$.009} & \textbf{.600{\scriptsize$\,\pm$.010}} & \textbf{.600{\scriptsize$\,\pm$.011}} & .625{\scriptsize$\,\pm$.006} \\
{\footnotesize $C{=}2$} & MLP AUC $\downarrow$ & .999{\scriptsize$\,\pm$.000} & .964{\scriptsize$\,\pm$.004} & .963{\scriptsize$\,\pm$.006} & .961{\scriptsize$\,\pm$.006} & \textbf{.871{\scriptsize$\,\pm$.010}} \\
 & R@1 $\uparrow$ & .106{\scriptsize$\,\pm$.003} & .656{\scriptsize$\,\pm$.011} & \textbf{.658{\scriptsize$\,\pm$.010}} & \textbf{.658{\scriptsize$\,\pm$.010}} & .559{\scriptsize$\,\pm$.012} \\
 & R@10 $\uparrow$ & .324{\scriptsize$\,\pm$.006} & .963{\scriptsize$\,\pm$.004} & \textbf{.964{\scriptsize$\,\pm$.004}} & \textbf{.964{\scriptsize$\,\pm$.004}} & .884{\scriptsize$\,\pm$.009} \\
\addlinespace[3pt]
\textbf{DE--IT} & Linear AUC $\downarrow$ & 1.000{\scriptsize$\,\pm$.000} & .613{\scriptsize$\,\pm$.018} & \textbf{.589{\scriptsize$\,\pm$.017}} & \textbf{.589{\scriptsize$\,\pm$.016}} & .604{\scriptsize$\,\pm$.016} \\
{\footnotesize $C{=}2$} & MLP AUC $\downarrow$ & 1.000{\scriptsize$\,\pm$.000} & .961{\scriptsize$\,\pm$.002} & .962{\scriptsize$\,\pm$.003} & .961{\scriptsize$\,\pm$.002} & \textbf{.870{\scriptsize$\,\pm$.005}} \\
 & R@1 $\uparrow$ & .054{\scriptsize$\,\pm$.001} & .647{\scriptsize$\,\pm$.010} & \textbf{.649{\scriptsize$\,\pm$.011}} & \textbf{.649{\scriptsize$\,\pm$.011}} & .536{\scriptsize$\,\pm$.002} \\
 & R@10 $\uparrow$ & .190{\scriptsize$\,\pm$.005} & \textbf{.957{\scriptsize$\,\pm$.004}} & \textbf{.957{\scriptsize$\,\pm$.004}} & \textbf{.957{\scriptsize$\,\pm$.005}} & .880{\scriptsize$\,\pm$.006} \\
\addlinespace[3pt]
\textbf{FR--IT} & Linear AUC $\downarrow$ & .999{\scriptsize$\,\pm$.001} & .611{\scriptsize$\,\pm$.016} & .577{\scriptsize$\,\pm$.015} & \textbf{.576{\scriptsize$\,\pm$.015}} & .606{\scriptsize$\,\pm$.016} \\
{\footnotesize $C{=}2$} & MLP AUC $\downarrow$ & .999{\scriptsize$\,\pm$.001} & .883{\scriptsize$\,\pm$.012} & .882{\scriptsize$\,\pm$.011} & .883{\scriptsize$\,\pm$.011} & \textbf{.764{\scriptsize$\,\pm$.017}} \\
 & R@1 $\uparrow$ & .722{\scriptsize$\,\pm$.010} & \textbf{.978{\scriptsize$\,\pm$.002}} & \textbf{.978{\scriptsize$\,\pm$.002}} & \textbf{.978{\scriptsize$\,\pm$.002}} & .965{\scriptsize$\,\pm$.006} \\
 & R@10 $\uparrow$ & .966{\scriptsize$\,\pm$.003} & \textbf{1.000{\scriptsize$\,\pm$.000}} & \textbf{1.000{\scriptsize$\,\pm$.000}} & \textbf{1.000{\scriptsize$\,\pm$.000}} & .999{\scriptsize$\,\pm$.000} \\
\addlinespace[3pt]
\textbf{DE--FR--IT} & Linear AUC $\downarrow$ & .999{\scriptsize$\,\pm$.000} & .906{\scriptsize$\,\pm$.004} & \textbf{.577{\scriptsize$\,\pm$.011}} & \textbf{.577{\scriptsize$\,\pm$.011}} & .615{\scriptsize$\,\pm$.010} \\
{\footnotesize $C{=}3$} & MLP AUC $\downarrow$ & .999{\scriptsize$\,\pm$.000} & .997{\scriptsize$\,\pm$.001} & .924{\scriptsize$\,\pm$.003} & .923{\scriptsize$\,\pm$.003} & \textbf{.838{\scriptsize$\,\pm$.008}} \\
 & R@1 $\uparrow$ & .294{\scriptsize$\,\pm$.004} & .585{\scriptsize$\,\pm$.006} & \textbf{.759{\scriptsize$\,\pm$.007}} & \textbf{.759{\scriptsize$\,\pm$.007}} & .688{\scriptsize$\,\pm$.004} \\
 & R@10 $\uparrow$ & .492{\scriptsize$\,\pm$.004} & .897{\scriptsize$\,\pm$.003} & \textbf{.969{\scriptsize$\,\pm$.003}} & \textbf{.969{\scriptsize$\,\pm$.003}} & .916{\scriptsize$\,\pm$.003} \\
\addlinespace[5pt]
\midrule
\addlinespace[4pt]
\multicolumn{7}{@{}l}{\textbf{Fedlex} \quad {\footnotesize confounder: language; 5 seeds (0--4), split by whole law}} \\
\cmidrule(lr){1-7}
\textbf{DE--FR--IT} & Linear AUC $\downarrow$ & 1.000{\scriptsize$\,\pm$.000} & .911{\scriptsize$\,\pm$.006} & \textbf{.644{\scriptsize$\,\pm$.016}} & \textbf{.644{\scriptsize$\,\pm$.016}} & .672{\scriptsize$\,\pm$.016} \\
{\footnotesize $C{=}3$} & MLP AUC $\downarrow$ & 1.000{\scriptsize$\,\pm$.000} & .990{\scriptsize$\,\pm$.001} & .958{\scriptsize$\,\pm$.003} & .958{\scriptsize$\,\pm$.002} & \textbf{.851{\scriptsize$\,\pm$.007}} \\
 & R@1 $\uparrow$ & .456{\scriptsize$\,\pm$.009} & .690{\scriptsize$\,\pm$.009} & \textbf{.804{\scriptsize$\,\pm$.008}} & \textbf{.804{\scriptsize$\,\pm$.008}} & .731{\scriptsize$\,\pm$.010} \\
 & R@10 $\uparrow$ & .784{\scriptsize$\,\pm$.009} & .950{\scriptsize$\,\pm$.004} & \textbf{.981{\scriptsize$\,\pm$.002}} & \textbf{.981{\scriptsize$\,\pm$.002}} & .938{\scriptsize$\,\pm$.005} \\
\addlinespace[5pt]
\midrule
\addlinespace[4pt]
\multicolumn{7}{@{}l}{\textbf{SCOTUS} \quad {\footnotesize confounder: source; 5 seeds (0--4), coverage cohort}} \\
\cmidrule(lr){1-7}
\textbf{Oyez--Lexis} & Linear AUC $\downarrow$ & 1.000{\scriptsize$\,\pm$.000} & .714{\scriptsize$\,\pm$.015} & .672{\scriptsize$\,\pm$.016} & \textbf{.622{\scriptsize$\,\pm$.017}} & .701{\scriptsize$\,\pm$.022} \\
{\footnotesize $C{=}2$} & MLP AUC $\downarrow$ & .998{\scriptsize$\,\pm$.000} & .892{\scriptsize$\,\pm$.010} & .890{\scriptsize$\,\pm$.010} & .874{\scriptsize$\,\pm$.009} & \textbf{.748{\scriptsize$\,\pm$.012}} \\
 & R@1 $\uparrow$ & .539{\scriptsize$\,\pm$.013} & .797{\scriptsize$\,\pm$.012} & .800{\scriptsize$\,\pm$.010} & \textbf{.801{\scriptsize$\,\pm$.010}} & .792{\scriptsize$\,\pm$.013} \\
 & R@10 $\uparrow$ & .811{\scriptsize$\,\pm$.014} & .942{\scriptsize$\,\pm$.004} & \textbf{.943{\scriptsize$\,\pm$.003}} & \textbf{.943{\scriptsize$\,\pm$.004}} & .931{\scriptsize$\,\pm$.002} \\
\addlinespace[3pt]
\textbf{Oyez--Wikipedia} & Linear AUC $\downarrow$ & 1.000{\scriptsize$\,\pm$.000} & .668{\scriptsize$\,\pm$.015} & \textbf{.633{\scriptsize$\,\pm$.005}} & \textbf{.633{\scriptsize$\,\pm$.010}} & .658{\scriptsize$\,\pm$.022} \\
{\footnotesize $C{=}2$} & MLP AUC $\downarrow$ & .999{\scriptsize$\,\pm$.000} & .855{\scriptsize$\,\pm$.017} & .855{\scriptsize$\,\pm$.010} & .853{\scriptsize$\,\pm$.009} & \textbf{.709{\scriptsize$\,\pm$.020}} \\
 & R@1 $\uparrow$ & .380{\scriptsize$\,\pm$.011} & .691{\scriptsize$\,\pm$.005} & \textbf{.694{\scriptsize$\,\pm$.006}} & \textbf{.694{\scriptsize$\,\pm$.005}} & .716{\scriptsize$\,\pm$.010} \\
 & R@10 $\uparrow$ & .652{\scriptsize$\,\pm$.005} & .915{\scriptsize$\,\pm$.007} & \textbf{.918{\scriptsize$\,\pm$.009}} & \textbf{.918{\scriptsize$\,\pm$.009}} & .917{\scriptsize$\,\pm$.007} \\
\addlinespace[3pt]
\textbf{Lexis--Wikipedia} & Linear AUC $\downarrow$ & .999{\scriptsize$\,\pm$.000} & .756{\scriptsize$\,\pm$.011} & .706{\scriptsize$\,\pm$.021} & \textbf{.648{\scriptsize$\,\pm$.015}} & .722{\scriptsize$\,\pm$.007} \\
{\footnotesize $C{=}2$} & MLP AUC $\downarrow$ & .997{\scriptsize$\,\pm$.001} & .864{\scriptsize$\,\pm$.018} & .847{\scriptsize$\,\pm$.013} & .833{\scriptsize$\,\pm$.008} & \textbf{.761{\scriptsize$\,\pm$.014}} \\
 & R@1 $\uparrow$ & .230{\scriptsize$\,\pm$.010} & .379{\scriptsize$\,\pm$.007} & \textbf{.380{\scriptsize$\,\pm$.008}} & .370{\scriptsize$\,\pm$.011} & .352{\scriptsize$\,\pm$.008} \\
 & R@10 $\uparrow$ & .505{\scriptsize$\,\pm$.011} & .680{\scriptsize$\,\pm$.009} & \textbf{.681{\scriptsize$\,\pm$.009}} & .680{\scriptsize$\,\pm$.009} & .635{\scriptsize$\,\pm$.015} \\
\addlinespace[3pt]
\textbf{All three sources} & Linear AUC $\downarrow$ & .999{\scriptsize$\,\pm$.000} & .882{\scriptsize$\,\pm$.004} & .699{\scriptsize$\,\pm$.085} & \textbf{.615{\scriptsize$\,\pm$.004}} & .727{\scriptsize$\,\pm$.009} \\
{\footnotesize $C{=}3$} & MLP AUC $\downarrow$ & .999{\scriptsize$\,\pm$.000} & \textbf{.979{\scriptsize$\,\pm$.000}} & .905{\scriptsize$\,\pm$.037} & .880{\scriptsize$\,\pm$.008} & .803{\scriptsize$\,\pm$.004} \\
 & R@1 $\uparrow$ & .350{\scriptsize$\,\pm$.006} & .506{\scriptsize$\,\pm$.007} & .560{\scriptsize$\,\pm$.033} & \textbf{.572{\scriptsize$\,\pm$.009}} & .563{\scriptsize$\,\pm$.007} \\
 & R@10 $\uparrow$ & .626{\scriptsize$\,\pm$.005} & .721{\scriptsize$\,\pm$.008} & .787{\scriptsize$\,\pm$.034} & \textbf{.800{\scriptsize$\,\pm$.008}} & .770{\scriptsize$\,\pm$.005} \\
\addlinespace[5pt]
\midrule
\addlinespace[4pt]
\multicolumn{7}{@{}l}{\textbf{Supplementary: other primary corpora} \quad {\footnotesize 5 seeds (0--4), primary cohort}} \\
\cmidrule(lr){1-7}
\textbf{SemEval} {\footnotesize $C{=}4$} & Graded sim.\ $\rho$ $\uparrow$ & .274{\scriptsize$\,\pm$.000} & .340{\scriptsize$\,\pm$.003} & .342{\scriptsize$\,\pm$.022} & .388{\scriptsize$\,\pm$.001} & \textbf{.401{\scriptsize$\,\pm$.003}} \\
\textbf{CAP} {\footnotesize $C{=}3$} & Topic ARI $\uparrow$ & .061{\scriptsize$\,\pm$.004} & .108{\scriptsize$\,\pm$.009} & .112{\scriptsize$\,\pm$.010} & \textbf{.146{\scriptsize$\,\pm$.010}} & .075{\scriptsize$\,\pm$.021} \\
\bottomrule
\end{tabular}
}
\noindent
\ifdim\wd\extendedIDsafebox>\dimexpr\linewidth-12pt\relax
  \resizebox{\dimexpr\linewidth-12pt\relax}{!}{\usebox{\extendedIDsafebox}}%
\else
  \usebox{\extendedIDsafebox}%
\fi
\par\vspace{6pt}
\refstepcounter{table}%
\label{tab:extended-id}%
\noindent{\raggedright\fontsize{10}{12}\selectfont\textbf{Table~\thetable:}\enspace
Extended in-distribution results for the eight E5-small composition settings. Cells report the mean $\pm$ sample SD across fitting seeds 0–4, using per-seed held-out scores without cross-seed pooling. Fedlex uses whole-law outer splits as in Table~\ref{tab:primary-id}.
\par}
\endgroup
\par

%% file: tables_4/tableB2_ten_encoder_robustness_singlepage.tex
\begin{table*}[p]
\centering

\fontsize{9.5pt}{11.5pt}\selectfont
\setlength{\tabcolsep}{1.0pt}
\renewcommand{\arraystretch}{1.07}
\def\robustTriple#1#2#3{\ensuremath{#1} & \ensuremath{#2} & \ensuremath{#3}}
\def\robustPair#1#2{\ensuremath{#1} & \ensuremath{#2}}

\vspace{3pt}
\noindent{\raggedright\textbf{Panel A.} $\Delta$ linear AUC (relative to Raw) \emph{(lower is better)}\par}
\par\vspace{3pt}
\noindent\resizebox{\dimexpr\linewidth-12pt\relax}{!}{%
\begin{tabular}{@{}l@{\hspace{6pt}}rrr@{\hspace{7pt}}rrr@{\hspace{7pt}}rrr@{\hspace{7pt}}rrr@{\hspace{7pt}}rrr@{}}
\toprule
 & \multicolumn{3}{c}{SemEval} & \multicolumn{3}{c}{SwiLTra DE--FR} & \multicolumn{3}{c}{SCOTUS O--L} & \multicolumn{3}{c}{CAP} & \multicolumn{3}{c}{Fedlex} \\
\cmidrule(lr){2-4}\cmidrule(lr){5-7}\cmidrule(lr){8-10}\cmidrule(lr){11-13}\cmidrule(l){14-16}
Encoder & PCA & KM & SAE & PCA & KM & SAE & PCA & KM & SAE & PCA & KM & SAE & PCA & KM & SAE \\
\midrule
MiniLM & \robustTriple{-.028}{-.323}{-.312} & \robustTriple{-.346}{-.379}{-.353} & \robustTriple{-.001}{.000}{-.127} & \robustTriple{-.077}{-.124}{-.275} & \robustTriple{-.088}{-.262}{-.269} \\
GIST-small & \robustTriple{-.008}{-.235}{-.283} & \robustTriple{-.333}{-.391}{-.370} & \robustTriple{-.001}{.000}{-.186} & \robustTriple{-.074}{-.117}{-.268} & \robustTriple{-.080}{-.182}{-.282} \\
GIST-base & \robustTriple{-.007}{-.123}{-.279} & \robustTriple{-.332}{-.363}{-.337} & \robustTriple{.000}{.000}{-.204} & \robustTriple{-.073}{-.107}{-.248} & \robustTriple{-.080}{-.298}{-.273} \\
GIST-large & \robustTriple{-.007}{-.182}{-.281} & \robustTriple{-.322}{-.351}{-.327} & \robustTriple{.000}{.000}{-.154} & \robustTriple{-.071}{-.106}{-.246} & \robustTriple{-.071}{-.278}{-.256} \\
E5-small & \robustTriple{-.070}{-.204}{-.303} & \robustTriple{-.371}{-.401}{-.377} & \robustTriple{-.276}{-.318}{-.283} & \robustTriple{-.113}{-.129}{-.296} & \robustTriple{-.091}{-.361}{-.329} \\
E5-base & \robustTriple{-.040}{-.228}{-.259} & \robustTriple{-.377}{-.389}{-.372} & \robustTriple{-.244}{-.310}{-.258} & \robustTriple{-.082}{-.114}{-.259} & \robustTriple{-.082}{-.111}{-.314} \\
E5-large & \robustTriple{-.085}{-.219}{-.275} & \robustTriple{-.348}{-.377}{-.345} & \robustTriple{-.174}{-.200}{-.234} & \robustTriple{-.050}{-.077}{-.235} & \robustTriple{-.069}{-.105}{-.308} \\
MPNet & \robustTriple{-.003}{-.152}{-.228} & \robustTriple{-.303}{-.340}{-.319} & \robustTriple{.000}{-.002}{-.143} & \robustTriple{-.076}{-.104}{-.242} & \robustTriple{-.081}{-.231}{-.239} \\
Nomic-v2 & \robustTriple{-.005}{-.005}{-.086} & \robustTriple{-.312}{-.383}{-.280} & \robustTriple{.000}{.000}{-.137} & \robustTriple{-.076}{-.109}{-.244} & \robustTriple{.000}{.000}{-.181} \\
MXB & \robustTriple{-.015}{-.263}{-.288} & \robustTriple{-.321}{-.345}{-.322} & \robustTriple{-.001}{.000}{-.183} & \robustTriple{-.077}{-.106}{-.253} & \robustTriple{-.069}{-.285}{-.260} \\
\midrule
\textbf{Mean} & \robustTriple{-.027}{-.194}{-.260} & \robustTriple{-.336}{-.372}{-.340} & \robustTriple{-.070}{-.083}{-.191} & \robustTriple{-.077}{-.109}{-.257} & \robustTriple{-.071}{-.211}{-.271} \\
\textbf{Range} & \robustTriple{.082}{.318}{.226} & \robustTriple{.075}{.061}{.096} & \robustTriple{.276}{.318}{.157} & \robustTriple{.063}{.052}{.061} & \robustTriple{.091}{.361}{.149} \\
\bottomrule
\end{tabular}
}

\par\vspace{6pt}
\noindent{\raggedright\textbf{Panel B.} $\Delta$ semantic utility (R@10; CAP: topic purity) \emph{(higher is better)}\par}
\par\vspace{3pt}
\noindent\resizebox{\dimexpr\linewidth-12pt\relax}{!}{%
\begin{tabular}{@{}l@{\hspace{6pt}}rrr@{\hspace{7pt}}rrr@{\hspace{7pt}}rrr@{\hspace{7pt}}rrr@{\hspace{7pt}}rrr@{}}
\toprule
 & \multicolumn{3}{c}{SemEval} & \multicolumn{3}{c}{SwiLTra DE--FR} & \multicolumn{3}{c}{SCOTUS O--L} & \multicolumn{3}{c}{CAP} & \multicolumn{3}{c}{Fedlex} \\
\cmidrule(lr){2-4}\cmidrule(lr){5-7}\cmidrule(lr){8-10}\cmidrule(lr){11-13}\cmidrule(l){14-16}
Encoder & PCA & KM & SAE & PCA & KM & SAE & PCA & KM & SAE & PCA & KM & SAE & PCA & KM & SAE \\
\midrule
MiniLM & \robustTriple{+.067}{+.243}{+.182} & \robustTriple{+.066}{+.065}{+.053} & \robustTriple{-.010}{-.010}{-.015} & \robustTriple{+.084}{+.119}{-.027} & \robustTriple{+.100}{+.155}{+.117} \\
GIST-small & \robustTriple{+.036}{+.100}{+.008} & \robustTriple{+.169}{+.173}{+.158} & \robustTriple{+.002}{+.001}{+.005} & \robustTriple{+.137}{+.146}{-.016} & \robustTriple{+.081}{+.125}{+.121} \\
GIST-base & \robustTriple{+.041}{+.093}{+.035} & \robustTriple{+.280}{+.278}{+.251} & \robustTriple{-.004}{-.005}{+.007} & \robustTriple{+.134}{+.153}{-.002} & \robustTriple{+.141}{+.303}{+.214} \\
GIST-large & \robustTriple{+.033}{+.088}{+.063} & \robustTriple{+.235}{+.237}{+.184} & \robustTriple{+.001}{.000}{+.005} & \robustTriple{+.156}{+.157}{-.018} & \robustTriple{+.162}{+.306}{+.216} \\
E5-small & \robustTriple{+.212}{+.217}{+.214} & \robustTriple{+.643}{+.645}{+.567} & \robustTriple{+.134}{+.135}{+.121} & \robustTriple{+.085}{+.085}{+.032} & \robustTriple{+.169}{+.201}{+.155} \\
E5-base & \robustTriple{+.039}{+.082}{+.050} & \robustTriple{+.381}{+.381}{+.361} & \robustTriple{+.158}{+.158}{+.148} & \robustTriple{+.096}{+.096}{+.034} & \robustTriple{+.040}{+.041}{+.033} \\
E5-large & \robustTriple{+.141}{+.166}{+.144} & \robustTriple{+.100}{+.100}{+.098} & \robustTriple{+.119}{+.121}{+.099} & \robustTriple{+.055}{+.077}{+.005} & \robustTriple{+.016}{+.017}{+.018} \\
MPNet & \robustTriple{+.015}{+.077}{+.064} & \robustTriple{+.097}{+.099}{+.064} & \robustTriple{+.002}{+.003}{+.007} & \robustTriple{+.114}{+.117}{-.034} & \robustTriple{+.106}{+.166}{+.122} \\
Nomic-v2 & \robustTriple{-.008}{-.005}{-.017} & \robustTriple{+.078}{+.079}{+.062} & \robustTriple{-.005}{-.002}{+.009} & \robustTriple{+.130}{+.126}{+.022} & \robustTriple{-.004}{-.004}{-.002} \\
MXB & \robustTriple{+.076}{+.197}{+.141} & \robustTriple{+.295}{+.294}{+.245} & \robustTriple{+.006}{+.001}{+.017} & \robustTriple{+.114}{+.149}{+.034} & \robustTriple{+.222}{+.397}{+.295} \\
\midrule
\textbf{Mean} & \robustTriple{+.065}{+.126}{+.088} & \robustTriple{+.234}{+.235}{+.204} & \robustTriple{+.040}{+.040}{+.040} & \robustTriple{+.110}{+.123}{+.003} & \robustTriple{+.103}{+.171}{+.129} \\
\bottomrule
\end{tabular}
}

\par\vspace{6pt}
\noindent{\raggedright\textbf{Panel C.} Post-hoc $k$-means ARI and PCA capture \emph{(ARI / PCA capture)}\par}
\par\vspace{3pt}
\noindent\resizebox{\dimexpr\linewidth-12pt\relax}{!}{%
\begin{tabular*}{\linewidth}{@{}l@{\extracolsep{\fill}}r@{\extracolsep{0pt}\hspace{6pt}}r@{\extracolsep{\fill}}r@{\extracolsep{0pt}\hspace{6pt}}r@{\extracolsep{\fill}}r@{\extracolsep{0pt}\hspace{6pt}}r@{\extracolsep{\fill}}r@{\extracolsep{0pt}\hspace{6pt}}r@{\extracolsep{\fill}}r@{\extracolsep{0pt}\hspace{6pt}}r@{}}
\toprule
 & \multicolumn{2}{c}{SemEval} & \multicolumn{2}{c}{\shortstack{SwiLTra\\DE--FR}} & \multicolumn{2}{c}{\shortstack{SCOTUS\\O--L}} & \multicolumn{2}{c}{CAP} & \multicolumn{2}{c}{Fedlex} \\
\cmidrule(lr){2-3}\cmidrule(lr){4-5}\cmidrule(lr){6-7}\cmidrule(lr){8-9}\cmidrule(l){10-11}
Encoder & ARI & PCA & ARI & PCA & ARI & PCA & ARI & PCA & ARI & PCA \\
\midrule
MiniLM & \robustPair{.55}{.78} & \robustPair{.99}{.99} & \robustPair{.00}{.05} & \robustPair{.59}{.89} & \robustPair{.84}{.91} \\
GIST-small & \robustPair{.37}{.71} & \robustPair{.99}{.99} & \robustPair{.00}{.09} & \robustPair{.61}{.92} & \robustPair{.71}{.91} \\
GIST-base & \robustPair{.33}{.72} & \robustPair{1.00}{1.00} & \robustPair{.01}{.09} & \robustPair{.65}{.95} & \robustPair{1.00}{.98} \\
GIST-large & \robustPair{.34}{.68} & \robustPair{.99}{1.00} & \robustPair{.01}{.09} & \robustPair{.62}{.96} & \robustPair{1.00}{.99} \\
E5-small & \robustPair{.77}{.93} & \robustPair{1.00}{1.00} & \robustPair{.84}{.98} & \robustPair{.68}{.97} & \robustPair{1.00}{.97} \\
E5-base & \robustPair{.76}{.79} & \robustPair{1.00}{1.00} & \robustPair{.88}{.98} & \robustPair{.68}{.93} & \robustPair{.57}{.54} \\
E5-large & \robustPair{.78}{.80} & \robustPair{1.00}{.99} & \robustPair{.64}{.94} & \robustPair{.60}{.94} & \robustPair{.55}{.49} \\
MPNet & \robustPair{.30}{.75} & \robustPair{.98}{.99} & \robustPair{.01}{.04} & \robustPair{.62}{.91} & \robustPair{.83}{.79} \\
Nomic-v2 & \robustPair{.01}{.20} & \robustPair{.98}{.97} & \robustPair{.00}{.03} & \robustPair{.68}{.94} & \robustPair{.00}{.01} \\
MXB & \robustPair{.52}{.70} & \robustPair{.99}{1.00} & \robustPair{.01}{.15} & \robustPair{.63}{.96} & \robustPair{1.00}{.99} \\
\bottomrule
\end{tabular*}%
}

\par\vspace{5pt}
\begin{minipage}{\dimexpr\linewidth-12pt\relax}
\end{minipage}

\caption{\textbf{Robustness across embedding models.} Each erasure method is refit in every encoder's embedding space (rather than transferred as a frozen transformation). Panels A and B report changes relative to Raw, averaged over matched runs using seeds 0--2. \textit{Notes.} Range = max--min; PCA = PCA-1; KM = $k$-means LEACE (silhouette-selected $K$); SAE = SAE-LEACE ($N=32$). The mean and range summarize ten encoders; $0$ denotes no change from Raw.}
\label{tab:ten-encoder-robustness}
\end{table*}

%% file: tables_4/table_SAE_construction_block.tex
\centering
\footnotesize
\setlength{\tabcolsep}{5pt}
\begin{tabular}{lcccrll}
\toprule
 &  &  &  & \multicolumn{3}{c}{Linear AUC$\downarrow$} \\
\cmidrule(lr){5-7}
TopK SAE & Eligible & Displ. & $r^{*}$ & MI-selected & Random $-$ MI & Matched PCA $-$ MI \\
\midrule
\multicolumn{7}{l}{\textbf{SwiLTra (DE/FR/IT)}} \\[1pt]
$8d$ latents, $k{=}32$ (the paper's) & 76--93 & .36 & 22 & .611{\scriptsize$\,\pm$.010} & +.015{\scriptsize\,[+.008, +.022]} & $-$.016{\scriptsize\,[$-$.025, $-$.008]} \\
$16d$ latents, $k{=}64$ & 303--351 & .35 & 22 & .604{\scriptsize$\,\pm$.006} & +.045{\scriptsize\,[+.020, +.070]} & $-$.009{\scriptsize\,[$-$.019, +.002]} \\
$8d$ latents, $k{=}64$ & 1470--1493 & .35 & 21--22 & .611{\scriptsize$\,\pm$.011} & +.063{\scriptsize\,[+.037, +.088]} & $-$.015{\scriptsize\,[$-$.019, $-$.012]} \\
$4d$ latents, $k{=}16$ & 42--56 & .36 & 23--24 & .604{\scriptsize$\,\pm$.011} & +.005{\scriptsize\,[.000, +.011]} & $-$.009{\scriptsize\,[$-$.025, +.008]} \\
\midrule
\multicolumn{7}{l}{\textbf{SCOTUS (3 sources)}} \\[1pt]
$8d$ latents, $k{=}32$ (the paper's) & 65--80 & .30 & 22 & .732{\scriptsize$\,\pm$.008} & +.010{\scriptsize\,[$-$.012, +.032]} & $-$.055{\scriptsize\,[$-$.079, $-$.031]} \\
$16d$ latents, $k{=}64$ & 287--318 & .30 & 22 & .715{\scriptsize$\,\pm$.003} & +.034{\scriptsize\,[+.007, +.060]} & $-$.038{\scriptsize\,[$-$.051, $-$.024]} \\
$8d$ latents, $k{=}64$ & 1479--1549 & .29 & 21 & .721{\scriptsize$\,\pm$.016} & +.059{\scriptsize\,[+.027, +.091]} & $-$.044{\scriptsize\,[$-$.091, +.003]} \\
$4d$ latents, $k{=}16$ & 22 & \multicolumn{5}{l}{\textit{not comparable at $N{=}32$: too few eligible latents (seed 0)}} \\
\bottomrule
\end{tabular}
\captionsetup{skip=2pt}
\captionof{table}{\textbf{SAE construction.}
E5-small, seeds 0--2, $N{=}32$. Rows vary TopK dictionary size and sparsity. \emph{Eligible}: latents with training firing rate in $[0.02,0.98]$; \emph{Displ.}: normalized squared training displacement of MI-selected SAE-LEACE; $r^{*}$: rank of displacement-matched PCA. Linear AUC is mean $\pm$ sample SD; random-SAE and matched-PCA columns report paired differences from MI selection with 95\% $t$-intervals. Settings with fewer than 32 eligible latents are not comparable.}
\label{tab:sae-construction}